\documentclass[10pt]{article} 

\usepackage[preprint]{rlj} 

\usepackage{amssymb}            
\usepackage{mathtools}          
\usepackage{mathrsfs}           
\usepackage{graphicx}           
\usepackage{subcaption}         
\usepackage[space]{grffile}     
\usepackage{url}                
\usepackage{lipsum}             

\usepackage{newfloat}
\usepackage{listings}
\usepackage{comment}
\usepackage{subcaption}
\usepackage{xcolor}
\usepackage{dashrule}
\usepackage{tabularx}
\definecolor{custompurple}{HTML}{800080}  
\definecolor{customblue}{HTML}{0000FF}  
\definecolor{customgreen}{HTML}{008000}  
\definecolor{custombrown}{HTML}{654321}  

\definecolor{color1}{rgb}{0.12, 0.47, 0.71} 
\definecolor{color2}{rgb}{0.17, 0.63, 0.17} 
\definecolor{color3}{rgb}{0.84, 0.15, 0.16} 
\definecolor{color4}{rgb}{0.58, 0.40, 0.74} 
\definecolor{color5}{rgb}{0.98, 0.94, 0.50} 
\definecolor{color6}{rgb}{0.53, 0.81, 0.98} 
\definecolor{color7}{rgb}{0.00, 0.50, 0.50} 
\definecolor{color8}{rgb}{1.00, 0.50, 0.31} 

\definecolor{dodgerblue}{HTML}{1E90FF}
\usepackage{pgfplots}
\pgfplotsset{compat=1.18}

\definecolor{rn6color}{RGB}{255,165,0}    
\definecolor{rn7color}{RGB}{119,221,119} 
\definecolor{rn4color}{RGB}{174,198,207} 
\definecolor{saccolor}{RGB}{25,25,112} 
\definecolor{bestcolor}{RGB}{0,120,200}   

\newcommand{\legendentry}[2]{%
  \textcolor{#1}{\rule{1.3em}{0.9ex}}\hspace{0.4em}\scriptsize #2%
}

\usepackage[capitalize,noabbrev]{cleveref}
\usepackage{placeins}
\usepackage{tikz}
\usepackage{pgfplots}
\usepackage{placeins} 
\usepgfplotslibrary{groupplots}

\usepgfplotslibrary{colorbrewer} 
\usepgfplotslibrary{external}    

\DeclareCaptionStyle{ruled}{labelfont=normalfont,labelsep=colon,strut=off} 
\usepackage{algorithm}
\usepackage{algorithmic}

\usepackage{amsmath}
\usepackage{amssymb}
\usepackage{float}    
\usepackage{placeins} 

\usepackage{newfloat}
\usepackage{listings}
\usepackage{comment}
\usepackage{subcaption}
\usepackage{xcolor}
\usepackage{dashrule}
\usepackage{tabularx}
\definecolor{custompurple}{HTML}{800080}  
\definecolor{customblue}{HTML}{0000FF}  
\definecolor{customgreen}{HTML}{008000}  
\definecolor{custombrown}{HTML}{654321}  

\usepackage{pgfplots}
\pgfplotsset{compat=1.18}
\usepackage{xcolor}
\usepackage{pgfplots}
\pgfplotsset{compat=1.18}
\usepgfplotslibrary{groupplots}
\usepackage{booktabs}    
\usepackage{array}  
\definecolor{dodgerblue}{HTML}{1E90FF}

\usepackage[capitalize,noabbrev]{cleveref}
\DeclareCaptionStyle{ruled}{labelfont=normalfont,labelsep=colon,strut=off} 
\floatstyle{ruled}
\newfloat{listing}{tb}{lst}{}
\floatname{listing}{Listing}

\title{Repetition as Reinforcement: Enhancing Sample Efficiency via Instant Episode Repetition in Reinforcement Learning}

\setrunningtitle{Instant Episode Repetition in Reinforcement Learning}

\title{Repetition as Reinforcement: Enhancing Sample Efficiency via Instant Episode Repetition in Reinforcement Learning}

\setrunningtitle{Instant Episode Repetition in Reinforcement Learning}

\author{
Hoda Yamani\textsuperscript{1},
Yuning Xing\textsuperscript{1},
Koen van Rijnsoever\textsuperscript{1},
Bruce A. MacDonald\textsuperscript{1},
Henry Williams\textsuperscript{1}
}

\coverPageAuthor{Hoda Yamani, Yuning Xing, Koen van Rijnsoever, Bruce A. MacDonald, Henry Williams}

\emails{
hoda.yamani@auckland.ac.nz,
yxin683@aucklanduni.ac.nz,
kvan910@aucklanduni.ac.nz,
b.macdonald@auckland.ac.nz,
henry.williams@auckland.ac.nz
}

\affiliations{
$^{1}$\textbf{Robot Learning Team, Center for Automation and Robotic Engineering Science (CARES),}\\
\textbf{Department of Electrical, Computer and Software Engineering, University of Auckland, New Zealand}
}

\contribution{
We introduce \textbf{IER}, a biologically inspired mechanism that modifies the RL interaction loop by immediately re-executing the action sequence from a prior high-reward episode.
}
{
Unlike replay-buffer methods that reuse experience only after storage, IER intervenes during data collection by replacing policy-selected actions with actions from a successful prior episode.
}

\contribution{
IER introduces episode-level behavioral consolidation by repeating complete action sequences from high-reward episodes during environment interaction, rather than treating experience as independent transition samples.
}
{
This shifts experience reuse from passive replay-buffer sampling toward active episode-level action replay, enabling successful behaviors to be revisited while new transitions are still being collected.
}

\contribution{
    We provide a simple and general integration of IER into off-policy continuous-control algorithms, demonstrating compatibility with SAC and TD3 without architectural modifications.
}
{
    The method operates at the interaction level and does not require changes to network structures or loss functions.
}

\contribution{
   Through evaluation on MuJoCo, the DeepMind Control Suite, and a real-world robotic manipulation task, we demonstrate that immediate episode repetition improves sample efficiency and accelerates convergence under consistent training settings.
}
{
    The empirical analysis compares IER against standard off-policy baselines and self-imitation methods under consistent training settings.
}


\summary{
Repetition is a central mechanism in biological learning, where revisiting successful experiences strengthens memory and stabilizes skill acquisition. Inspired by this principle, we propose \textbf{Instant Episode Repetition (IER)}, a simple mechanism that improves sample efficiency in reinforcement learning by actively re-executing action sequences from high-reward episodes during environment interaction. Unlike replay-buffer methods, which passively reuse stored transitions during policy updates, IER directly modifies the data collection process: when an episode achieves a higher cumulative reward than previously observed, the agent immediately re-executes the same action sequence under different initial conditions for a fixed number of repetitions. IER integrates seamlessly with off-policy continuous-control algorithms, including SAC and TD3, without architectural modifications. Experiments on MuJoCo, the DeepMind Control Suite, and a real-world robotic manipulation task show that IER improves learning efficiency and accelerates policy refinement compared with standard off-policy and SIL-enhanced baselines, suggesting that structured action-sequence repetition provides a simple yet effective mechanism for enhancing reinforcement learning.

}

\begin{document}

\makeCover  
\maketitle  

\begin{abstract}
Repetition is a fundamental mechanism in human learning, where revisiting successful experiences strengthens memory, consolidates skills, and improves future performance. Motivated by this biological principle, we introduce Instant Episode Repetition (IER), a simple and novel mechanism that improves sample efficiency by immediately repeating action sequences from successful episodes during environment interaction. Unlike conventional approaches such as Experience Replay and Self-Imitation Learning (SIL), which passively reuse past experience during training updates, IER directly influences the data collection process. Upon identifying a high-reward episode, the agent repeats its action sequence for a fixed number of subsequent episodes, reinforcing valuable behaviors through renewed interaction with the environment. We integrate IER into state-of-the-art SAC and TD3 algorithms and evaluate its effectiveness on continuous-control benchmarks, including MuJoCo, the DeepMind Control Suite, and a real-world dynamic object translation task with a robotic manipulator. Experimental results demonstrate that this simple mechanism improves learning performance over standard and self-imitation-based baselines.
\end{abstract}

\section{Introduction}
Reinforcement Learning (RL) offers a powerful framework for training agents to make sequential decisions by interacting with their environment and learning from feedback  \cite{sutton1998reinforcement}. 
RL has achieved notable success in various domains, from robotic manipulation and autonomous driving to game playing \cite{han2023survey, silver2016mastering, kiran2021deep}. However, one of the key challenges that continues to limit its broader applicability is sample efficiency, the ability to learn effective policies from a limited number of interactions \cite{ding2020challenges}. In contrast to biological agents, which can often learn complex behaviors from relatively few experiences, RL agents typically require millions of interactions to achieve comparable performance \cite{lake2017building,bellemare2016unifying}. This inefficiency presents a significant barrier, especially in real-world environments where interactions are expensive and time-consuming \cite{dulac2021challenges}.

Neuro-scientific research highlights reward-driven repetition as a fundamental mechanism underlying efficient learning \cite{schultz2006behavioral, doya2008modulators}. When humans or animals experience a rewarding outcome, they are more likely to reengage in the same sequence of actions that led to that reward \cite{kahana2024laws}. This action repetition reinforces the link between actions and outcomes and supports the development of procedural memory and motor skills \cite{hartley2008learning, graybiel2008habits}. For example, when learning to ride a bike, a person may initially struggle, trying different ways to balance, steer, and pedal. However, once they successfully ride a short distance, they instinctively repeat the same coordination of movements \cite{ericsson1993role, kahana2024laws}. Each repetition improves their competence, and the successful action pattern becomes more stable and efficient. At the neural level, this process strengthens synaptic connections through mechanisms such as synaptic plasticity and dopaminergic modulation, which support the consolidation and refinement of effective strategies over time \cite{hebb2005organization, schultz1997neural}.

\begin{figure}[!t]
    \centering
    \includegraphics[width=0.99\textwidth]{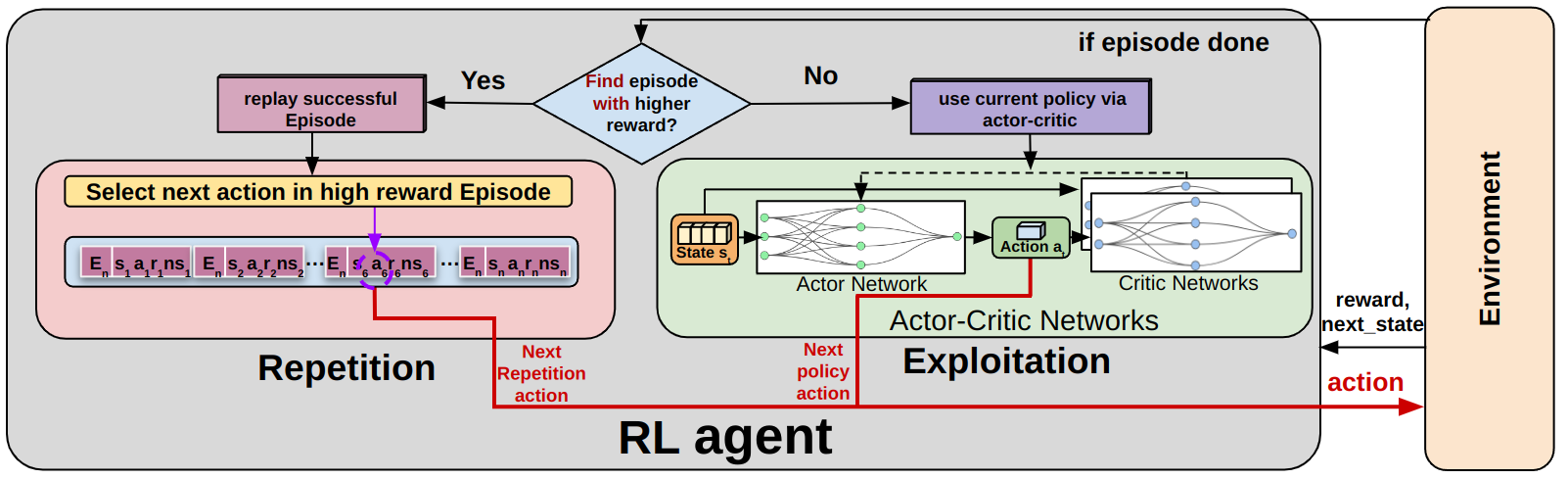}
    \caption[High-level overview of the Instant Episode Repetition algorithm]{Illustration of the Instant Episode Repetition Algorithm: At the end of each episode (when done is true), if the agent identifies an \textbf{episode with a high reward}, it immediately initiates repetition by executing the actions from that episode sequentially, from start to end, for \(RN\) repetitions within the environment.}

    \label{fig:instant-repetition}
\end{figure}

Inspired by this, we investigate how repetition can be explicitly incorporated into RL frameworks to reinforce high-value experiences and improve learning. While existing techniques such as experience replay \cite{lin1992self} and PER \cite{schaul2015prioritized} enhance training efficiency by reusing stored transitions, they do so passively during policy updates and do not directly influence how the agent behaves during interaction. Consequently, even successful behaviors are rarely repeated in the environment, delaying their reinforcement.

Recent methods, such as Self-Imitation Learning (SIL) \cite{oh2018self}, build on this idea by encouraging agents to imitate past experiences with high cumulative rewards obtained within an episode. SIL samples episodes from the replay buffer and prioritizes high returns to guide learning. However, these approaches remain passive and do not directly alter the agent’s interaction with the environment \cite{dai2020episodic}.

We present \textbf{Instant Episode Repetition (IER)}, a novel and conceptually simple mechanism designed to improve sample efficiency in RL by reinforcing successful behaviors immediately upon discovery. In contrast to conventional off-policy approaches, which execute policy-generated actions during interaction and rely primarily on passive reuse of stored transitions in replay buffers, IER modifies the data collection process through repetition of action sequences from newly discovered high-reward episodes. When the agent identifies an episode achieving a new highest reward, it stores the corresponding action sequence and repeats it immediately after the current episode for a fixed number of subsequent episodes, instead of sampling actions from the current policy. By generating additional samples around behaviors that have already demonstrated high value, IER increases the density of informative experience in promising regions of the state--action space. Importantly, this intervention affects only the interaction strategy and does not alter the underlying network architectures, objective functions, or optimization procedures.

We integrate IER into standard off-policy frameworks and evaluate its effectiveness across continuous-control benchmarks in both simulated and real-world robotic environments. The experimental results demonstrate that IER outperforms the baseline algorithms, highlighting its potential to improve learning efficiency and performance.~\textbf{Code availability:} The implementation of IER is publicly available at \url{https://github.com/UoA-CARES/instant-episode-repetition}.


\section{Background and Related Work}

\subsection{Behavioral and Neural Foundations of Repetition}

Repetition and reinforcement are fundamental mechanisms of adaptive learning in biological systems~\cite{haleem2015role}. Reinforcement increases the likelihood of repeating actions associated with rewarding outcomes, a principle formalized in operant conditioning and supported by dopaminergic reward prediction signaling~\cite{thorndike2017animal, schultz1997neural}. Repetition, in turn, stabilizes and refines reinforced behaviors through repeated activation of neural circuits and structured practice, strengthening synaptic efficacy and supporting long-term memory consolidation~\cite{hebb2005organization, bliss1973long}.  

At the neural level, repeated activation of task-relevant circuits strengthens synaptic connections, a process captured by Hebbian learning~\cite{hebb2005organization}. Long-term potentiation provides physiological evidence that sustained co-activation enhances synaptic efficacy and supports memory consolidation~\cite{bliss1973long}. Through these mechanisms, repetition transforms transient successes into stable behavioral patterns and progressively increases execution efficiency~\cite{graybiel2008habits}.

Importantly, repetition is not mechanical duplication. Behavioral and motor learning research emphasizes “repetition without repetition,” where each execution occurs under slightly varying conditions, promoting robustness and generalization~\cite{bernstein1967coordination, latash2012bliss}. Neuroimaging studies further demonstrate repetition suppression, reflecting increased processing efficiency as familiar patterns are re-engaged~\cite{henson2003neuroimaging}. Together, these findings indicate that effective learning arises from repeatedly re-engaging successful behaviors under natural variability.

These principles suggest that repetition consolidates success by stabilizing effective behaviors while preserving adaptability. Translating this insight to RL motivates the deliberate re-execution of high-reward episodes as a means of reinforcing valuable strategies during interaction. Motivated by this biological perspective, we introduce IER as an interaction-level repetition mechanism for policy learning.

\subsection{Self-Imitation Learning (SIL)}
\label{sec:SIL}

SIL is a value-based RL technique that improves policy learning by leveraging past experiences with high discounted return~\cite{oh2018self}. It performs off-policy updates by increasing the likelihood of actions whose discounted returns exceed the current value estimate. However, SIL remains a passive method: although it updates the policy based on successful past episodes, the agent still samples actions from its current policy, and high-reward trajectories do not directly shape new experience. This separation between learning and behavior limits SIL’s ability to reinforce successful strategies during interaction.

Recent extensions of SIL  improve learning efficiency by integrating additional learning signals or auxiliary mechanisms. For example,~\cite{li2023accelerating} present A-SILfD, which incorporates expert demonstrations and constrains updates using ensemble Q-functions to mitigate overestimation. However, their approach relies on expert data emphasizing key behaviors, limiting its applicability where such data is unavailable or difficult to obtain.
 
\cite{andres2022towards} combine SIL with intrinsic motivation to guide exploration toward previously successful behaviors, with intrinsic signals further emphasizing the prioritization of high-reward episodes.~\cite{dai2020episodic} introduce an episodic-level replay strategy combined with hindsight to improve policy learning. While useful in goal-conditioned settings, relying on Hindsight Experience Replay (HER) limits its use in unstructured or general-purpose environments.
\cite{kuang2025goal} incorporate adversarial training to improve robustness within the SIL framework. However, their method operates primarily through offline replay and does not influence action selection during interaction, reducing its effectiveness for real-time adaptation and behavior shaping.

In contrast, the proposed IER approach moves beyond purely policy-driven sampling by allowing the agent to re-execute entire successful trajectories, thereby embedding the recall of past successes directly into the interaction process. By reinforcing temporally coherent action sequences rather than isolated transitions, IER strengthens valuable behavioral patterns while enabling on-policy evaluation under dynamically evolving conditions. To our knowledge, this is a novel direction in self-imitation that can be integrated into a wide range of algorithms, including both value-based and actor-critic methods, without requiring structural modifications or external demonstrations. 


\section{Methodology}

In this section, we introduce IER and describe how it integrates into standard off-policy actor--critic RL. IER operates at episode boundaries and alters the interaction dynamics while preserving the underlying learning objective and optimization procedure.

\subsection{Preliminaries}

RL is formulated as a Markov Decision Process (MDP)
\begin{equation}
\mathcal{M} = (\mathcal{S}, \mathcal{A}, P, R, \gamma),
\label{eq:mdp}
\end{equation}
where $\mathcal{S}$ and $\mathcal{A}$ denote the state and action spaces, 
$P(s' \mid s,a)$ is the transition kernel, 
$R(s,a)$ the reward function, 
and $\gamma \in [0,1)$ the discount factor.

An episode (trajectory) is defined as
\begin{equation}
\tau = \{(s_0,a_0,r_0), \ldots, (s_T,a_T,r_T)\},
\label{eq:trajectory}
\end{equation}
representing a complete interaction sequence from an initial state to termination.

In off-policy actor--critic methods such as TD3 and SAC, transitions $(s_t,a_t,r_t,s_{t+1})$ are stored in a replay buffer $\mathcal{M}$ and sampled to update the actor $\pi_\theta$ and critic $Q$ networks through temporal-difference learning.

For episode-level selection in IER, we define the episode reward as the sum of rewards accumulated over an episode,
\begin{equation}
R_{\text{ep}}(\tau) = \sum_{t=0}^{T} r_t.
\label{eq:episode_reward}
\end{equation}

IER uses this $R_{\text{ep}}(\tau)$ as the criterion for determining whether an episode should be repeated.

\subsection{Instant Episode Repetition}

IER modifies the data collection phase of off-policy RL by temporarily reinforcing high-reward episodes. It does not alter the reward function, loss formulation, network architecture, or optimization procedure. Instead, it modifies how episodes are generated.

At the end of each episode (when the \texttt{done} flag is true), the agent computes the episode reward $R_{\text{ep}}$ and updates the best observed value:
\begin{equation}
r_{\max} \leftarrow \max(r_{\max}, R_{\text{ep}}).
\label{eq:rmax_update}
\end{equation}

For the subsequent episode, the agent operates in one of two modes, as illustrated in Figure~\ref{fig:instant-repetition}:

\begin{itemize}

\item 
\textbf{Repetition:} 
If the current episode achieves a new maximum episode reward, the full action sequence
\begin{equation}
\mathbf{a}^\star = (a_0^\star, \dots, a_T^\star)
\label{eq:stored_sequence}
\end{equation}
is stored. The agent then enters a repetition phase lasting $RN$ consecutive episodes, where $RN$ is a hyperparameter controlling repetition strength. During this phase, the stored action sequence $\mathbf{a}^\star$ is executed sequentially instead of sampling from the policy.

\item 
\textbf{Exploitation:} 
If no new maximum-reward episode is detected, or once repetition ends, the agent resumes standard behavior by sampling actions from its current policy:
\begin{equation}
a_t \sim \pi_\theta(\cdot \mid s_t).
\label{eq:policy_sampling}
\end{equation}

\end{itemize}

All transitions collected during both repetition and exploitation are inserted into the replay buffer $\mathcal{M}$ and used for standard off-policy updates. Consequently, IER modifies only the visitation distribution of newly collected episodes while preserving the underlying learning dynamics. The complete training procedure is summarized in pseudo-code in the supplementary material.

\subsection{Re-execution Under New Initial Conditions}

A key property of IER is that repetition does not duplicate identical episodes. Although several benchmark environments are deterministic in their transition dynamics, each episode starts from an initial state sampled from a reset distribution $p_0$. As a result, re-executing the same stored action sequence $\mathbf{a}^\star$ does not reproduce the original episode exactly. Instead, it generates a family of locally related episodes in the neighborhood of the previously observed high-reward behavior.

Rather than cloning past experience, IER replays successful action sequences under perturbed initial conditions. While the action sequence remains fixed, the resulting trajectories differ at the state level due to variations in the starting state and subsequent state evolution. Consequently, repetition increases sampling density in regions of the state space that have empirically yielded high rewards, without duplicating identical transitions. This concentrates visitation around informative regions while avoiding collapse to a single deterministic trajectory.

In stochastic environments or real-world robotic systems, additional variability from sensor noise, contact dynamics, and actuation uncertainty further broadens this local sampling distribution. Thus, IER preserves replay diversity while reinforcing behaviorally salient regions, supporting stable and generalizable learning.


\section{Experiments}

We evaluate the proposed IER method on a diverse set of high-dimensional continuous control tasks in both simulated and real-world environments. The goal is to examine how episodic repetition affects learning performance when combined with state-of-the-art RL algorithms. The following subsections introduce the baseline algorithms used for comparison and describe the simulated and real-world tasks employed in the experiments.

\subsection{Baseline Algorithms and Variants}

We integrate IER into two widely used off-policy RL algorithms: TD3~\cite{fujimoto2018addressing} and SAC~\cite{haarnoja2018soft}, yielding the variants \textbf{IER-TD3} and \textbf{IER-SAC}. To isolate the effect of IER, we compare these agents with:
\begin{itemize}
    \item \textbf{TD3 and SAC:} Standard implementations with no episode repetition.
   \item \textbf{SIL-TD3} and \textbf{SIL-SAC}: We integrate SIL \cite{oh2018self} with TD3 following the implementation in \cite{hu2021generalizable}, and extend the same formulation to SAC for a fair comparison. We deliberately use the standard SIL formulation rather than newer variants to avoid introducing additional mechanisms (e.g., demonstrations, intrinsic rewards) discussed in Background and Related Works section, which could confound the comparison. This provides a clean baseline to highlight the behavioral difference between passive experience reuse in SIL and active episodic repetition in IER.
\end{itemize}

\subsection{Simulation Environments}

We evaluated IER and baseline algorithms on eight continuous control tasks spanning both the \textbf{DeepMind Control Suite}~\cite{tassa2018deepmind} and \textbf{MuJoCo}~\cite{todorov2012mujoco}. These simulation platforms are widely recognized benchmarks for RL, providing diverse dynamics, varying levels of task complexity, and high-dimensional state-action spaces that closely resemble real-world continuous control challenges. By selecting tasks from both environments, we ensure a comprehensive evaluation of IER's robustness and generalization across different domains and difficulty levels. Detailed descriptions and configurations of the specific tasks used in our experiments are provided in the supplementary material.


\subsection{Real-World Task: Dynamic Object Translation via In-Hand Manipulation}
\label{sec:real-world-eval}

\begin{figure}[t]
    \centering
    \includegraphics[
        width=0.45\textwidth,
        height=0.35\textheight,
        keepaspectratio
    ]{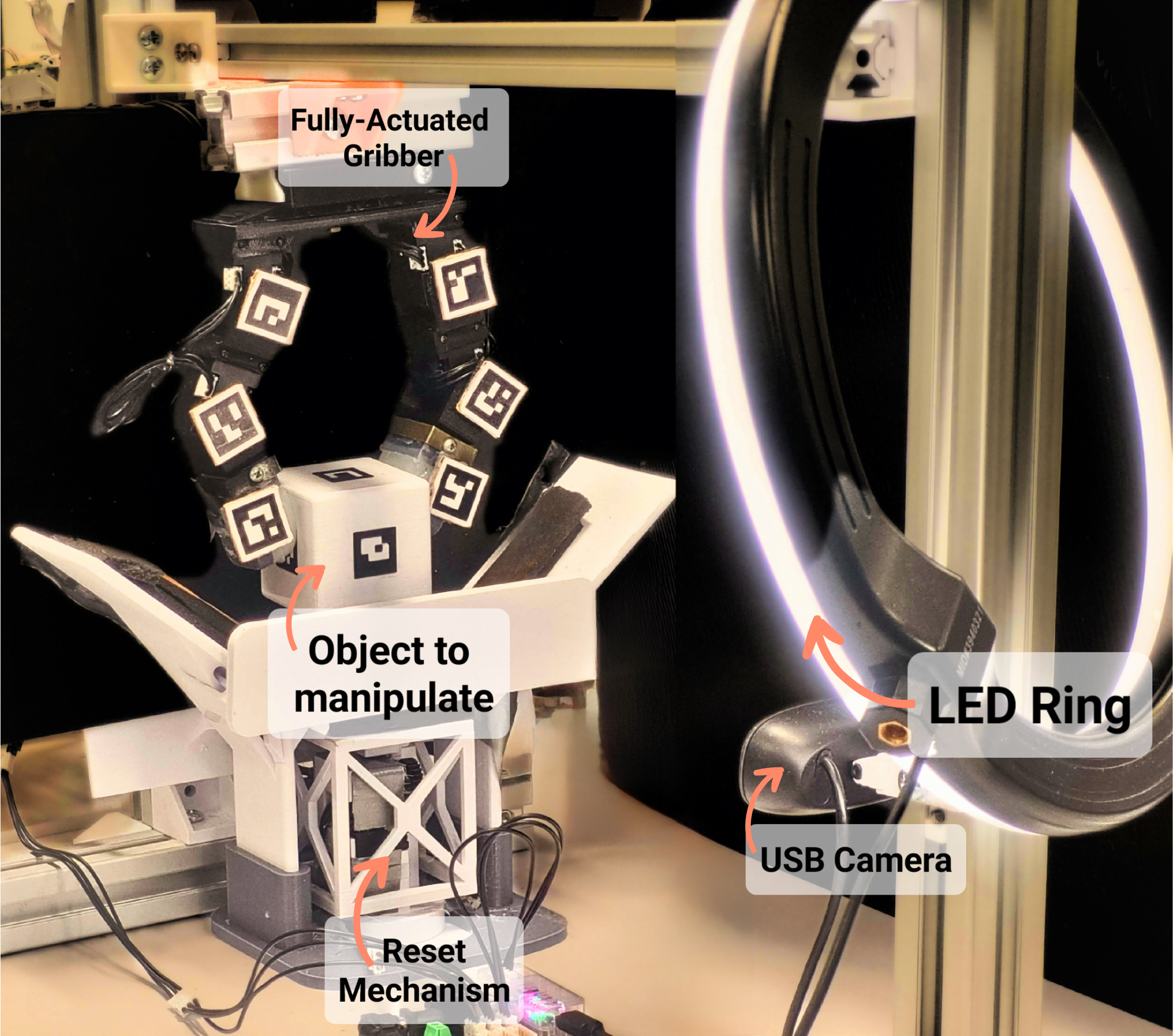}
    \caption{Real-world setup for dynamic object translation. The cube must be moved laterally while suspended in the air by a custom-built gripper.}
    \label{fig:gripper-setup}
\end{figure}

We evaluated our method on a stochastic dexterous in-hand manipulation task that requires moving a cube sideways while maintaining a stable grasp. As shown in Figure~\ref{fig:gripper-setup}, each episode starts with a reset mechanism that autonomously repositions the cube to ensure consistent trial conditions and minimize human intervention. The gripper must perform precise and coordinated movements to keep the cube from slipping or falling, relying entirely on its dexterity without external support.

To provide a reliable perception, the workspace is uniformly illuminated by a ring of LEDs, while a camera tracks the pose of the cube in real time using ArUco markers. The agent receives rewards for maintaining a stable grasp and increasing lateral displacement between time steps, encouraging fine-grained control while preserving grasp integrity. This setup captures key real-world challenges, including contact dynamics, gravitational effects, sensor noise, and constraints on data collection, while supporting scalable, physically grounded experimentation. Additional details on the hardware, reward function, and implementation are provided in the supplementary material.


\begin{figure}[t]
\centering
\captionsetup{font=small}
\def\imgw{0.245\linewidth}

\vspace{2pt}
{\large \textbf{SAC}}
\vspace{4pt}

\begin{subfigure}[b]{\imgw}
    \includegraphics[width=\linewidth]{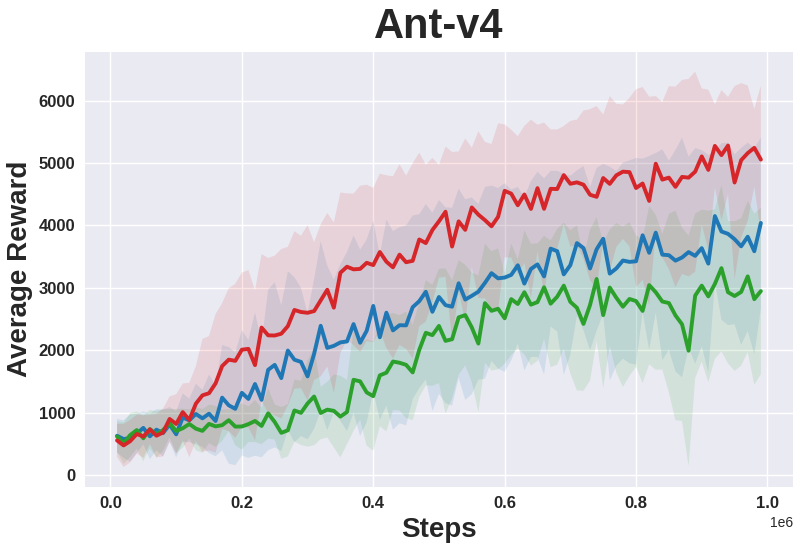}
\end{subfigure}\hfill
\begin{subfigure}[b]{\imgw}
    \includegraphics[width=\linewidth]{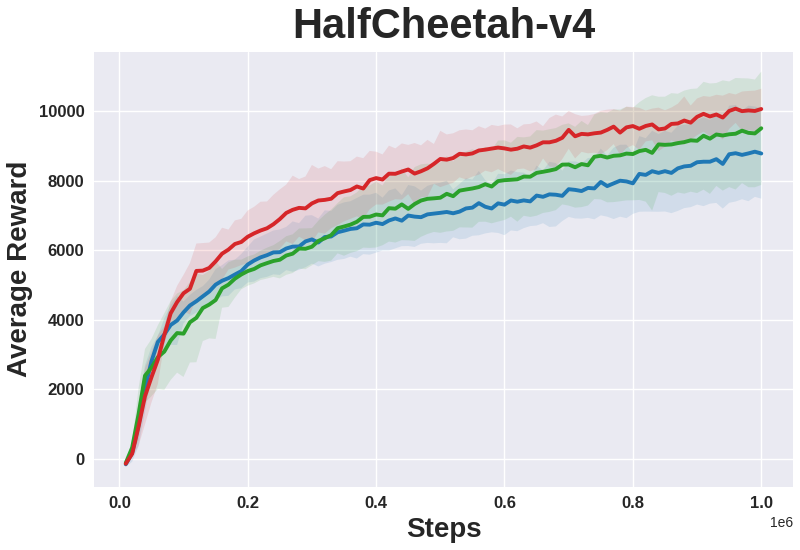}
\end{subfigure}\hfill
\begin{subfigure}[b]{\imgw}
    \includegraphics[width=\linewidth]{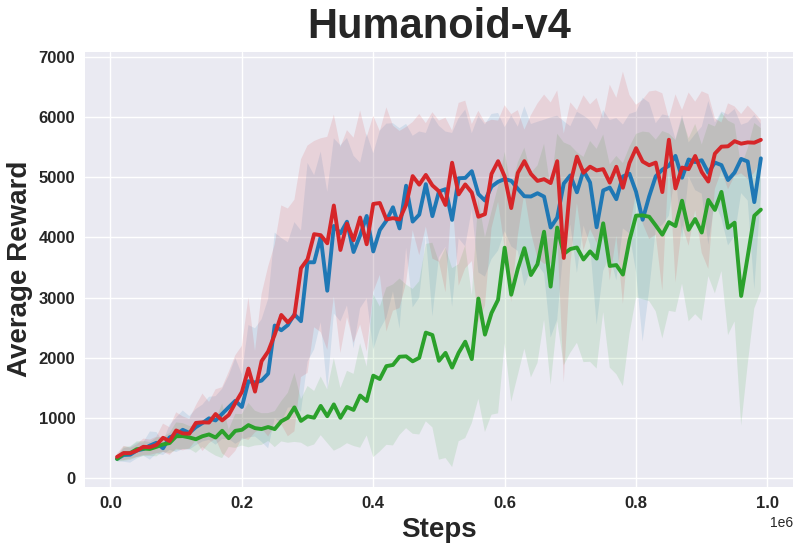}
\end{subfigure}\hfill
\begin{subfigure}[b]{\imgw}
    \includegraphics[width=\linewidth]{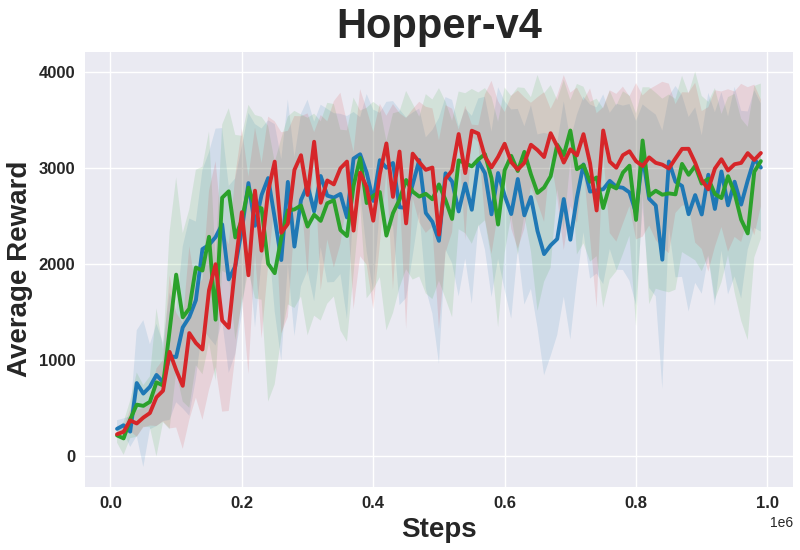}
\end{subfigure}

\vspace{2pt}

\begin{subfigure}[b]{\imgw}
    \includegraphics[width=\linewidth]{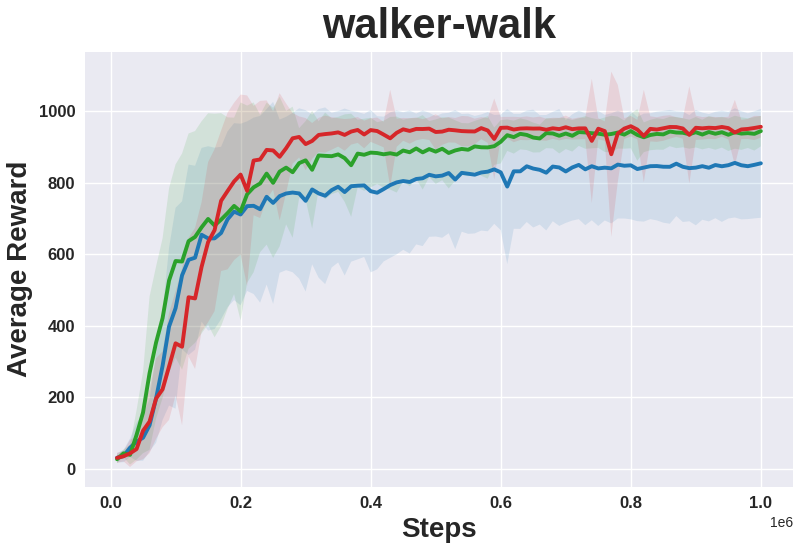}
\end{subfigure}\hfill
\begin{subfigure}[b]{\imgw}
    \includegraphics[width=\linewidth]{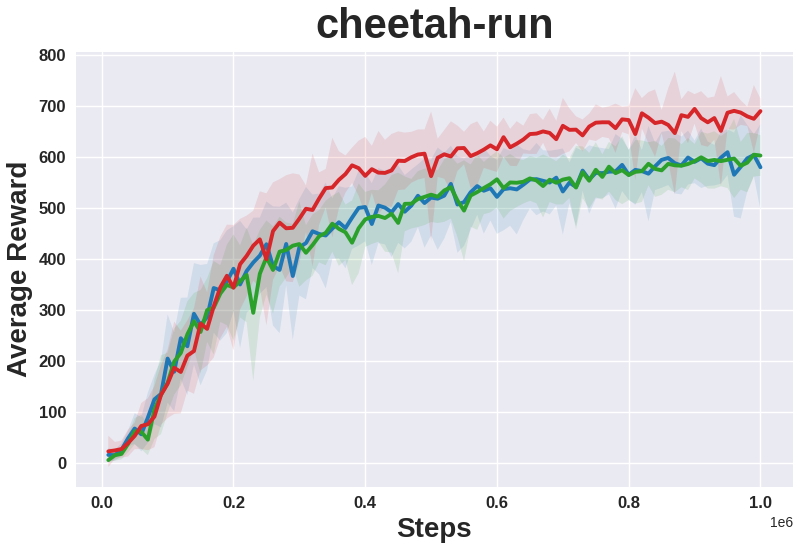}
\end{subfigure}\hfill
\begin{subfigure}[b]{\imgw}
    \includegraphics[width=\linewidth]{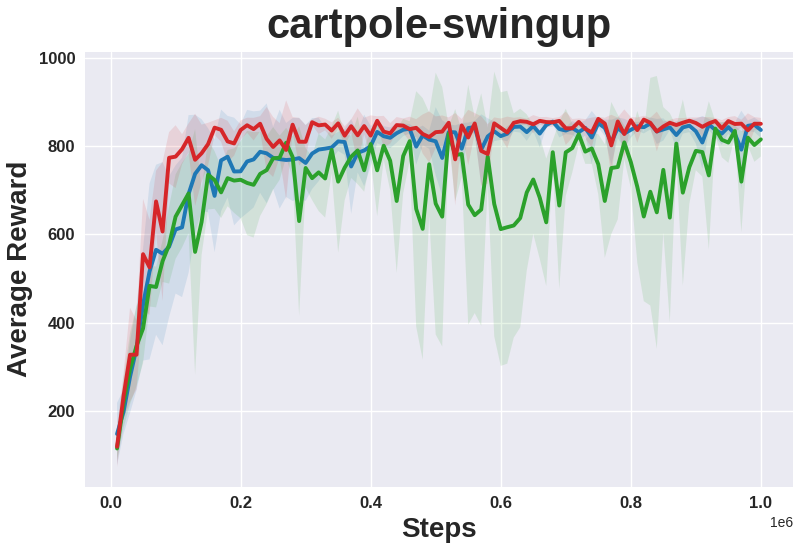}
\end{subfigure}\hfill
\begin{subfigure}[b]{\imgw}
    \includegraphics[width=\linewidth]{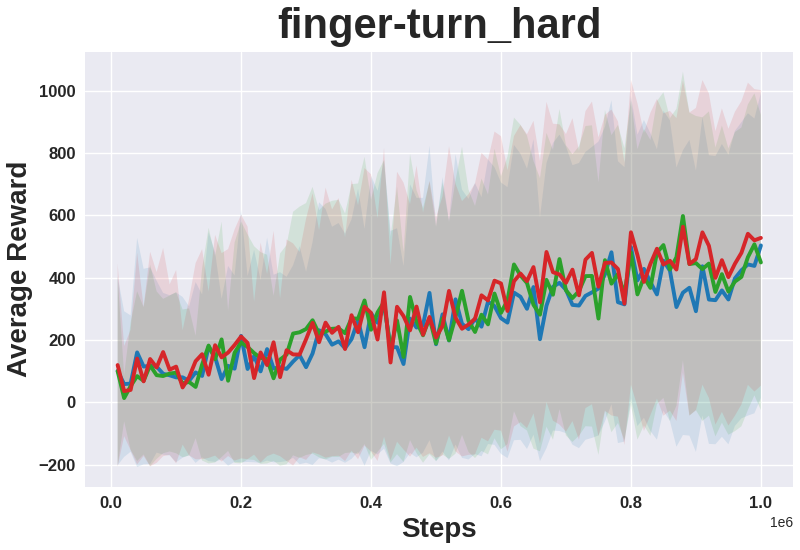}
\end{subfigure}

\vspace{10pt}
{\large \textbf{TD3}}
\vspace{4pt}

\begin{subfigure}[b]{\imgw}
    \includegraphics[width=\linewidth]{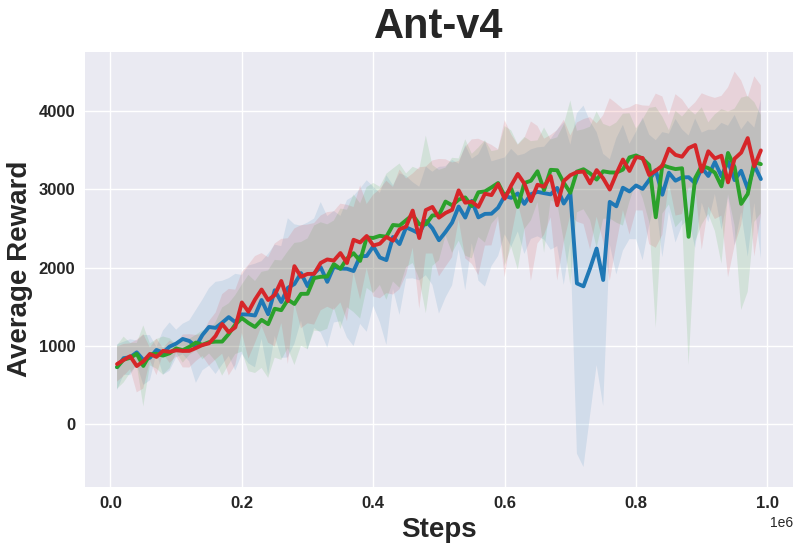}
\end{subfigure}\hfill
\begin{subfigure}[b]{\imgw}
    \includegraphics[width=\linewidth]{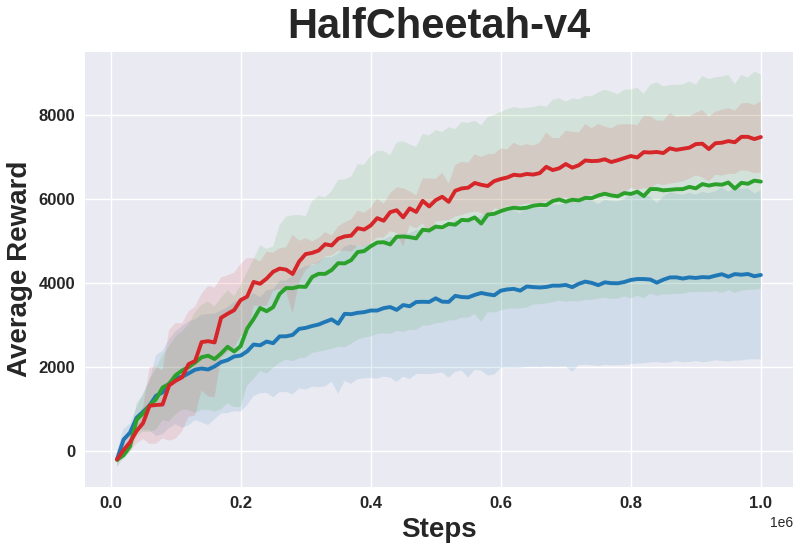}
\end{subfigure}\hfill
\begin{subfigure}[b]{\imgw}
    \includegraphics[width=\linewidth]{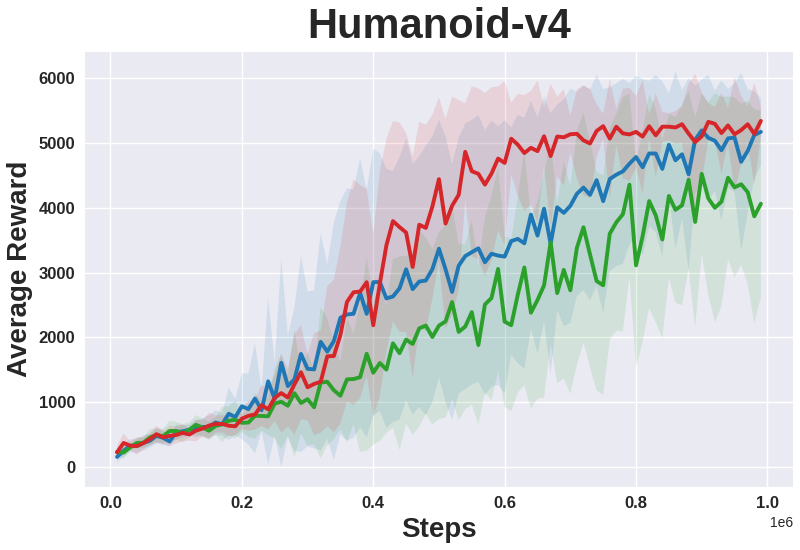}
\end{subfigure}\hfill
\begin{subfigure}[b]{\imgw}
    \includegraphics[width=\linewidth]{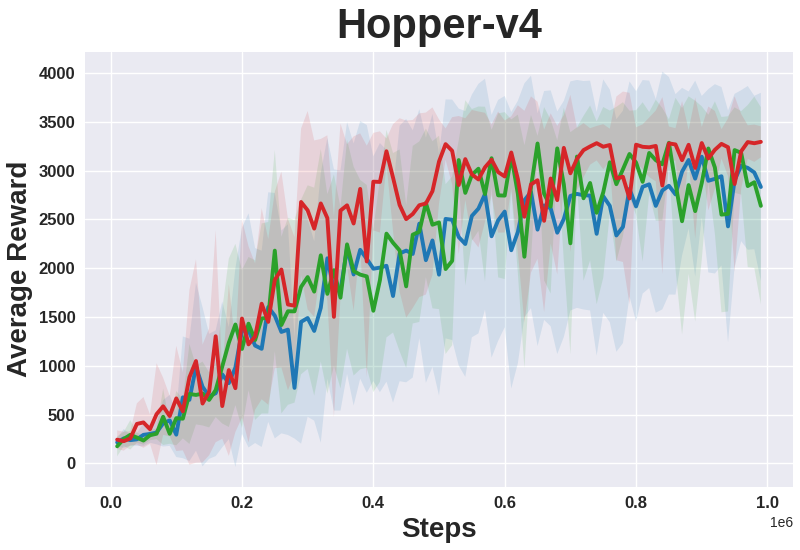}
\end{subfigure}

\vspace{2pt}

\begin{subfigure}[b]{\imgw}
    \includegraphics[width=\linewidth]{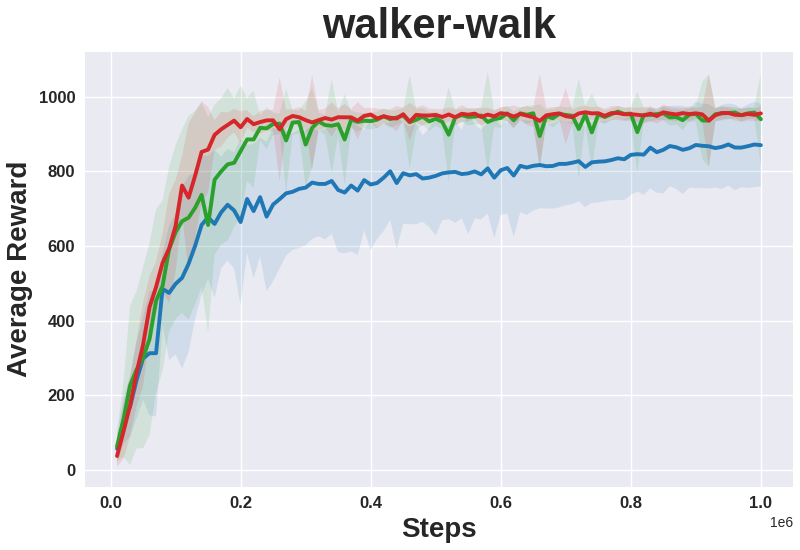}
\end{subfigure}\hfill
\begin{subfigure}[b]{\imgw}
    \includegraphics[width=\linewidth]{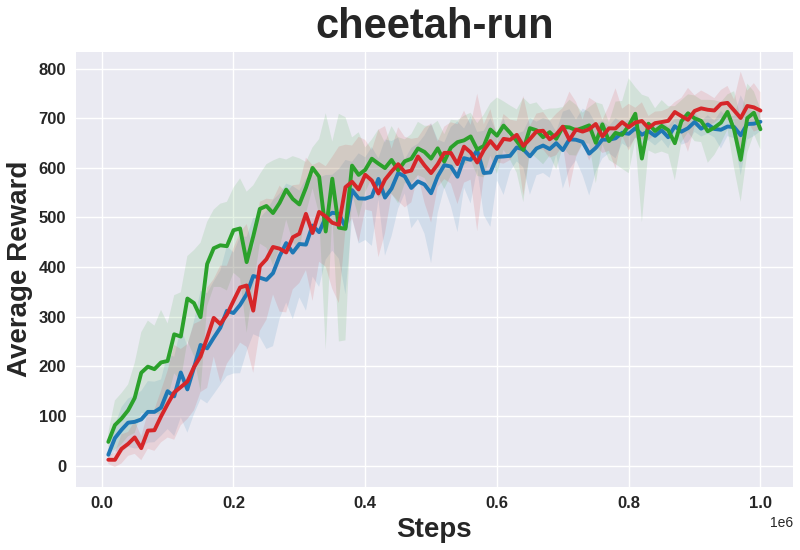}
\end{subfigure}\hfill
\begin{subfigure}[b]{\imgw}
    \includegraphics[width=\linewidth]{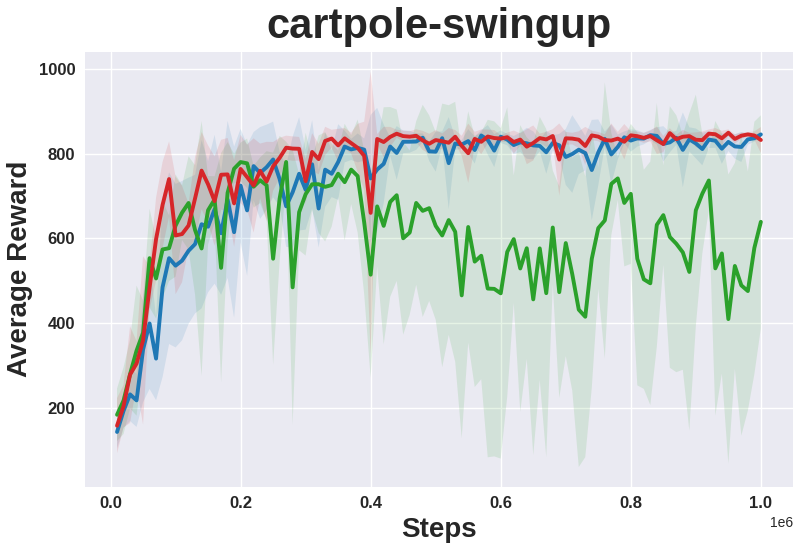}
\end{subfigure}\hfill
\begin{subfigure}[b]{\imgw}
    \includegraphics[width=\linewidth]{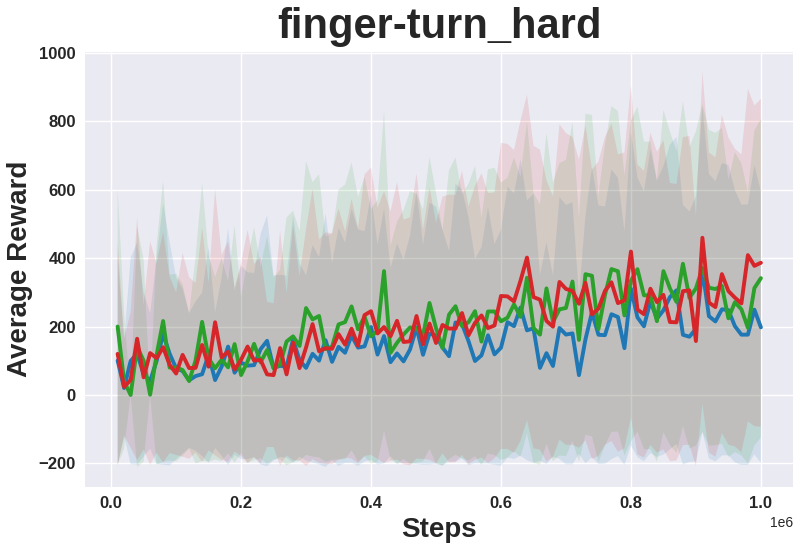}
\end{subfigure}

\vspace{6pt}
\begin{tabular}{@{}l@{}}
\textcolor{color1}{\rule{0.03\linewidth}{0.18cm}} \scriptsize Baseline \hspace{10pt}
\textcolor{color2}{\rule{0.03\linewidth}{0.18cm}} \scriptsize SIL \hspace{10pt}
\textcolor{color3}{\rule{0.03\linewidth}{0.18cm}} \scriptsize IER
\end{tabular}

\caption{
Performance comparison of Baseline, SIL, and the proposed IER method across eight continuous-control tasks under SAC (top) and TD3 (bottom). Curves show the mean episodic return over five independent seeds, and shaded regions represent the standard deviation across five random seeds. 
}
\label{fig:IER-all}
\end{figure}

    
 \section{Experimental Results}
 
   We evaluate the performance of the proposed repetition-based methods, \textbf{IER-SAC} and \textbf{IER-TD3}, across eight simulated continuous-control tasks and a real-world robotic task. These variants are compared against the standard SAC and TD3 baselines, along with their respective self-imitation learning variants (SIL-SAC and SIL-TD3). The learning curves for simulated tasks are shown in Figure~\ref{fig:IER-all}, which reports SAC-based methods (top) and TD3-based methods (bottom). Results for the real-world setup are presented in Figure~\ref{fig:gripper-combined}.

    Across most tasks, \textbf{IER-SAC} and \textbf{IER-TD3} consistently outperform their respective baselines, demonstrating that repetition of high-reward episodes improves learning efficiency and convergence speed with minimal additional complexity.
    
    In some environments, baseline performance is slightly lower than values reported in the original publications, likely due to simulator stochasticity and random seed variation. However, relative performance differences between competing methods remain consistent across seeds. Our evaluation follows established best practices for fair RL benchmarking, as outlined by \cite{henderson2018deep}.
    
   We begin by presenting results from simulated environments for both SAC- and TD3-based variants, followed by evaluation in a real-world setting. Finally, we conduct a sensitivity analysis to examine the impact of RN.

    \subsection{Performance in Simulated Environments}
    
    As shown in Figure~\ref{fig:IER-all}, \textbf{IER-SAC} improves over the SAC baseline in six out of eight tasks and performs comparably in the remaining two. The strongest gains occur in dynamic locomotion tasks such as \textit{Ant-v4}, \textit{HalfCheetah-v4}, \textit{Walker-Walk}, and \textit{Cheetah-Run}, where agents must learn coordinated and temporally extended movement patterns. In these domains, repetition reinforces stable behavioral sequences, leading to faster convergence and improved long-term rewards.
    
    \textbf{SIL-SAC} performs competitively in selected environments, including \textit{HalfCheetah-v4} and \textit{Walker-Walk}, but shows inconsistent performance in more unstable tasks such as \textit{Humanoid-v4} and \textit{Cartpole-Swingup}. In some cases, performance falls below the SAC baseline, reflecting a known limitation of self-imitation learning: the risk of reinforcing suboptimal trajectories when exploration remains incomplete~\cite{oh2018self, guo2019self}.
    
    Under TD3 (bottom row of Figure~\ref{fig:IER-all}), \textbf{IER-TD3} outperforms standard TD3 in six out of eight tasks and exceeds SIL-TD3 in five tasks, with notable improvements in \textit{HalfCheetah-v4}, \textit{Humanoid-v4}, and \textit{Walker-Walk}. These results further confirm that repetition strengthens beneficial trajectory segments that contribute to improved cumulative reward.
    
    While both IER variants outperform their respective baselines, the relative improvements observed with \textbf{IER-TD3} are generally smaller than those seen with \textbf{IER-SAC}. This difference may stem from the underlying exploration mechanisms. SAC incorporates entropy regularization, promoting broader exploration and maintaining policy diversity. The repetition mechanism complements this objective by reinforcing effective behaviors without substantially limiting exploration.
    
    In contrast, TD3 employs deterministic policy updates, which may be more sensitive to repeated behavior sequences and thus more prone to premature convergence if repetition is excessive. This architectural difference likely explains the comparatively smaller but still consistent gains observed for \textbf{IER-TD3}.

\subsection{Real-World Evaluation}
\label{sec:gripper-evaluation}

\begin{figure}[t]
    \centering

    \begin{subfigure}{0.49\linewidth}
        \centering
        \includegraphics[width=\linewidth]{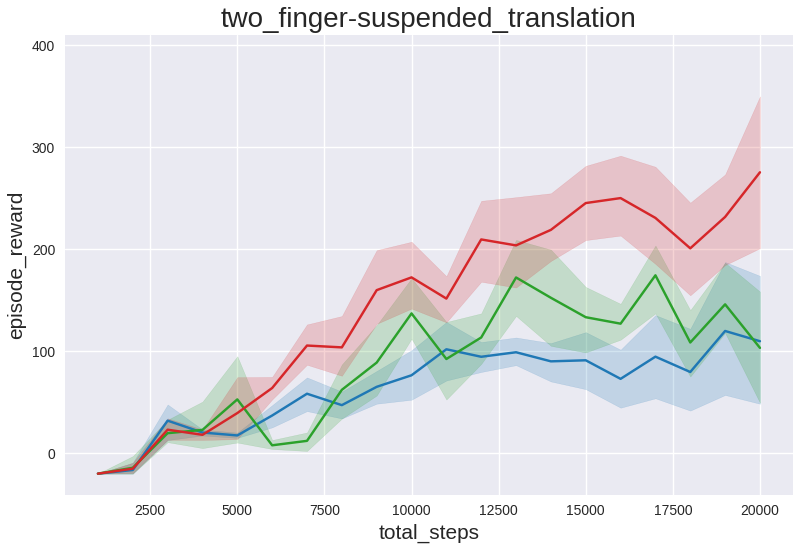}
        \caption{SAC variants}
        \label{fig:sac-gripper}
    \end{subfigure}
    \hfill
    \begin{subfigure}{0.49\linewidth}
        \centering
        \includegraphics[width=\linewidth]{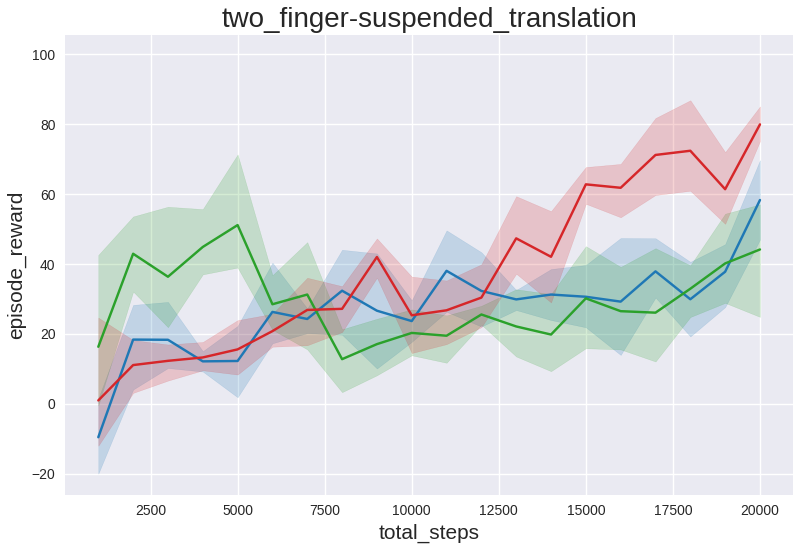}
        \caption{TD3 variants}
        \label{fig:td3-gripper}
    \end{subfigure}

    \vspace{4pt}

    \centering
    \begin{tabular}{@{}l@{\hspace{12pt}}l@{\hspace{12pt}}l@{}}
        \legendentry{color1}{Base} &
        \legendentry{color2}{SIL} &
        \legendentry{color3}{IER}
    \end{tabular}

    \vspace{-6pt}
    \caption{Performance comparison of standard, SIL-enhanced, and IER-enhanced SAC and TD3 variants in the real-world gripper manipulation task. Curves show the mean episodic return over five independent seeds, and shaded regions represent the standard deviation across five random seeds.}
    \label{fig:gripper-combined}
\end{figure}

To evaluate the robustness of the proposed IER method under real-world stochasticity, we conduct experiments on a physical robotic manipulation task (see Experiments Section). Unlike simulated benchmarks, real-world environments introduce unavoidable sources of variability, including sensor noise, friction and contact inconsistencies, and actuation delays. These factors make trajectory execution inherently stochastic, even when identical action commands are applied.

As shown in Figure~\ref{fig:gripper-combined}, \textbf{IER-SAC} achieves faster convergence and higher final performance compared to both SAC and SIL-SAC. A similar pattern is observed in the TD3-based setting, where \textbf{IER-TD3} outperforms TD3 and SIL-TD3, although with narrower margins. While SIL-TD3 exhibits a temporary early improvement, its performance quickly saturates and fails to sustain gains, allowing both TD3 and IER-TD3 to surpass it after relatively few environment interactions.

These results indicate that the repetition strategy remains effective under real-world stochasticity. In particular, IER improves data efficiency and promotes more stable learning dynamics in this setting.


\subsection{Sensitivity Analysis of the Repetition Number (RN)}

This section examines how the RN affects IER performance. RN specifies how many consecutive episodes re-execute a high-reward episode once it is identified. We evaluate RN values from $\{0,1,\ldots,7\}$ for both IER-SAC and IER-TD3 across all benchmark tasks.
Performance is measured using the normalized Area Under the Learning Curve (AUC), capturing both learning speed and stability. We report the relative improvement over RN=0 (no repetition):

\begin{equation}
\Delta \mathrm{AUC}\% =
\frac{\mathrm{AUC}_{\mathrm{RN}} - \mathrm{AUC}_{\mathrm{RN0}}}
{\mathrm{AUC}_{\mathrm{RN0}}}
\times 100.
\end{equation}

\subsubsection{Aggregate Trends Across Tasks}

\begin{figure}[t]
    \centering
    \includegraphics[width=0.45\textwidth]{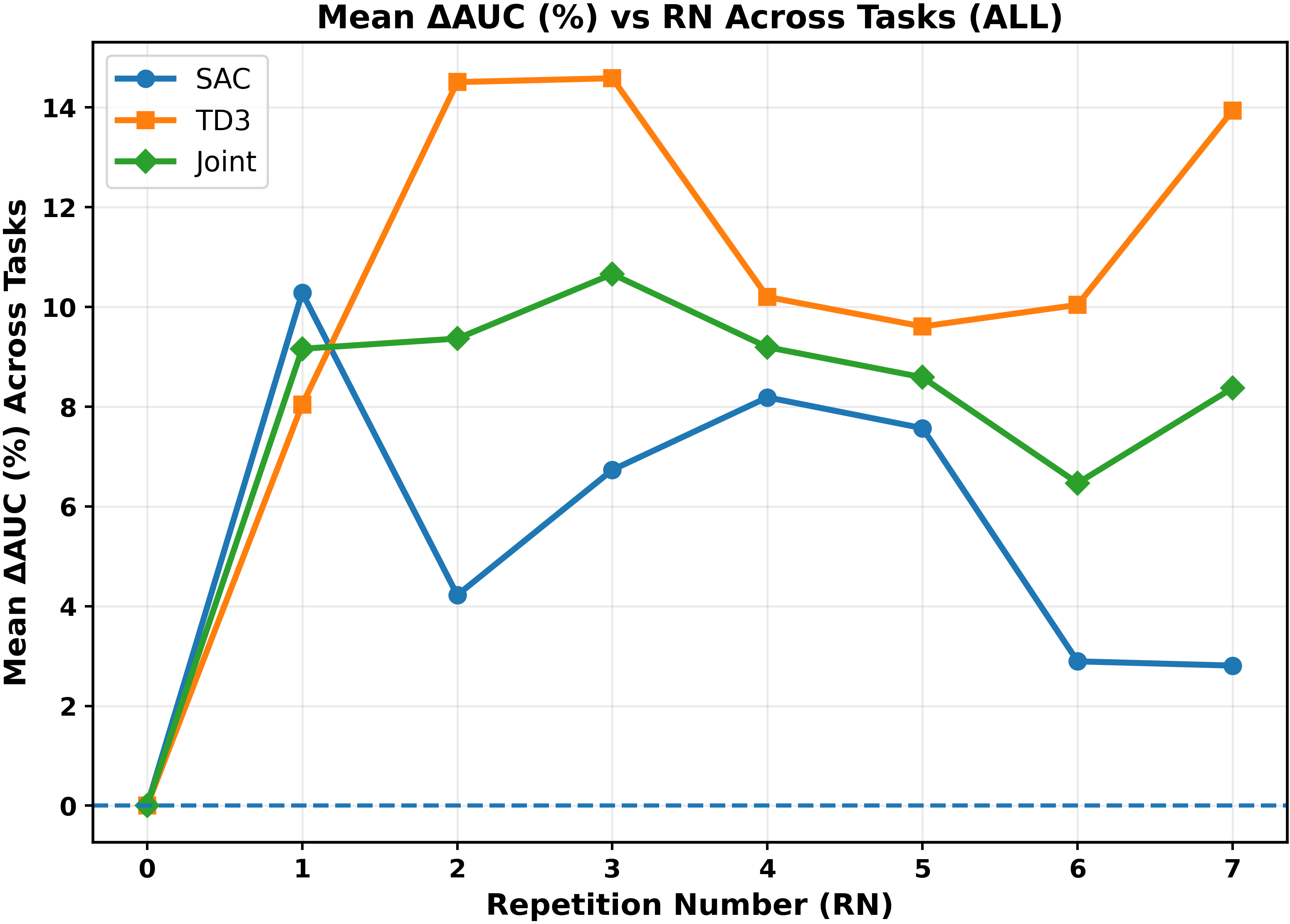}
    \caption{
    Mean $\Delta$AUC\% across all eight tasks as a function of the RN, jointly averaged over SAC and TD3.
    }
    \label{fig:mean_delta_curve_all}
\end{figure}

Figure~\ref{fig:mean_delta_curve_all} shows the mean $\Delta$AUC\% across all eight tasks as a function of RN, jointly averaged over SAC and TD3. Performance peaks at RN3 (10.66\%), followed by RN2 and RN4, and declines at larger repetition levels. The resulting unimodal pattern indicates that moderate repetition yields the strongest overall acceleration of learning. Small RN values may insufficiently reinforce high-reward episodes, whereas excessively large RN values can reduce state diversity, leading to diminishing returns.

We report the joint average across SAC and TD3 to provide a unified view of repetition effects across baseline algorithms. Although task-wise analyses are presented separately, the aggregate curve highlights repetition dynamics that remain consistent across settings and provides a compact estimate of the most effective repetition range.

\subsubsection{Task-Specific Trends and Algorithm Sensitivity}

\begin{figure}[t]
    \centering
    \includegraphics[width=0.98\textwidth]{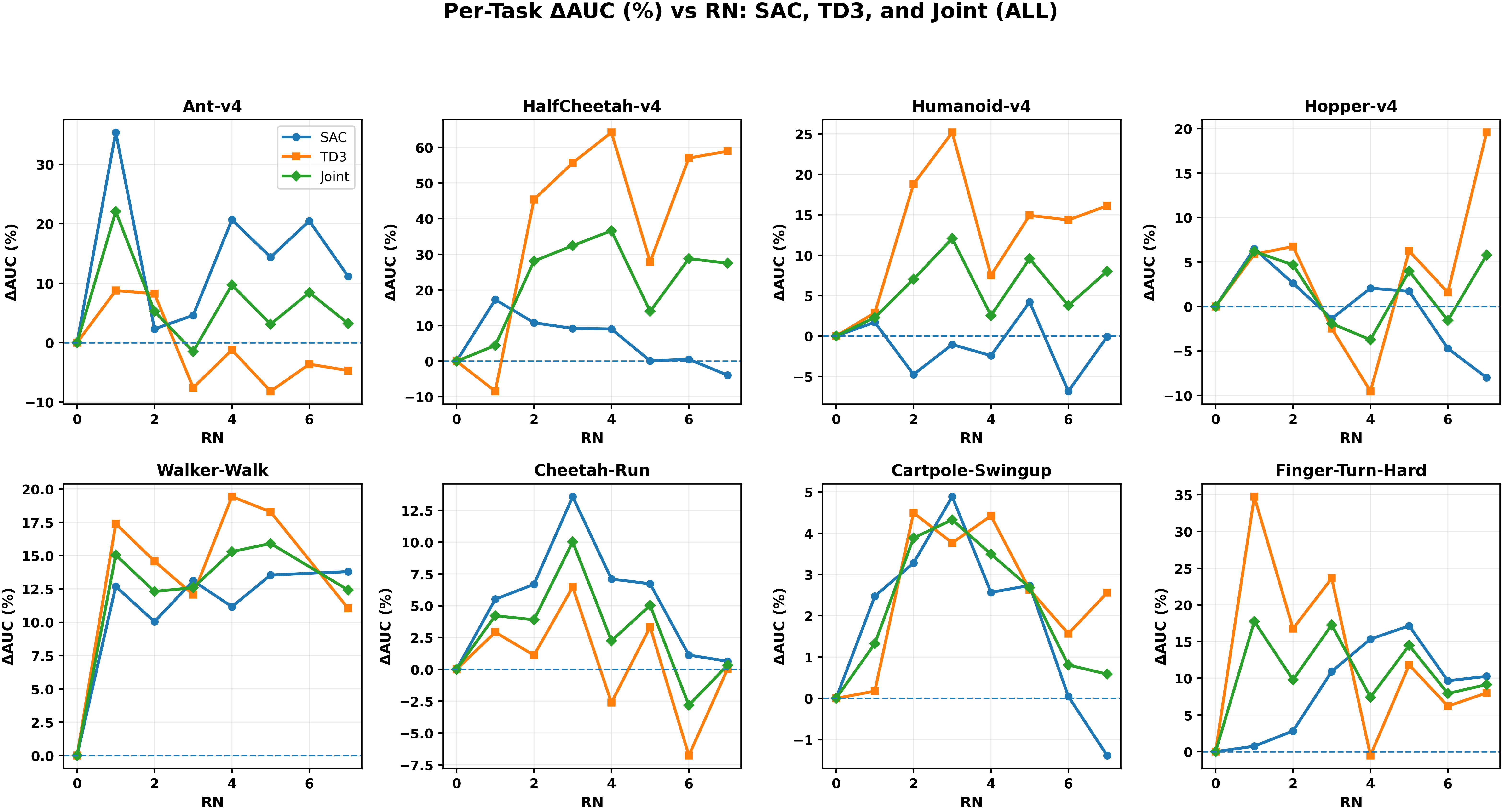}
    \caption{
    Per-task $\Delta$AUC\% as a function of the Repetition Number (RN), from RN0 (no repetition) to RN7, for SAC and TD3.
    }
    \label{fig:delta_auc_per_task}
\end{figure}

Figure~\ref{fig:delta_auc_per_task} presents the per-task $\Delta$AUC\% as a function of the RN for both TD3 and SAC. Results suggest that environments with higher instability or reward variance tend to benefit more strongly from repetition. For example, \textbf{Finger-Turn-Hard} and \textbf{Walker-Walk} show substantial gains at moderate or higher RN values. In \textbf{Walker-Walk} and \textbf{Cheetah-Run}, performance often improves with increasing RN and typically peaks at intermediate levels (RN3--RN5). In contrast, tasks such as \textbf{Hopper-v4} and \textbf{Cartpole-Swingup} exhibit smaller and more variable changes, consistent with their comparatively stable reward dynamics.

The impact of repetition varies across baseline algorithms. For TD3, performance commonly peaks at moderate RN values (RN2–RN3), yielding consistent improvements across multiple tasks. In contrast, SAC tends to benefit from smaller RN values, with performance exhibiting greater sensitivity at larger repetition levels. This difference reflects their respective update mechanisms: TD3 gains from stronger reinforcement of high-reward episodes, while SAC’s entropy regularization inherently promotes exploration, increasing its sensitivity to excessive repetition.

Although the optimal RN varies across environments and occasional performance declines are observed at certain repetition levels, moderate repetition (RN2--RN4) often achieves competitive and near-peak performance. These results indicate that RN should be treated as a tunable hyperparameter rather than a fixed setting.

Additional analyses, including complete RN0–RN7 performance curves, domain-level comparisons, and agreement analysis between TD3 and SAC, are provided in the supplementary material. These extended results further demonstrate the robustness of moderate repetition and confirm its effectiveness across both MuJoCo and DeepMind Control environments.


\section{\textbf{Conclusion}}

This work introduced IER, a repetition-based mechanism that enhances RL by directly reinforcing action sequences from high-reward episodes through immediate re-execution. Unlike traditional approaches that rely solely on replayed transitions, IER intervenes during data collection, enabling agents to revisit successful behaviours during data collection.

IER provides a simple, general, and biologically inspired extension to off-policy RL that requires no architectural modifications. Empirical results across simulated benchmarks and real-world robotic tasks demonstrate consistent improvements in data efficiency and training stability.

\bibliography{main}
\bibliographystyle{rlj}

\beginSupplementaryMaterials

This supplementary material provides additional methodological details and experimental results that complement the main paper. The content is organized as follows:

\begin{itemize}
    \item \textbf{Appendix A} – Pseudo-code of the proposed IER algorithm.
    \item \textbf{Appendix B} – Extended Analysis of the Repetition Number (RN)
    \item \textbf{Appendix C} – Description of the experimental environments, including MuJoCo tasks, the DeepMind Control Suite, and the real-world \textit{Dynamic Object Translation with Gripper Manipulation} setup.
    \item \textbf{Appendix D} – Detailed hyperparameter configurations for IER and baseline algorithms.
\end{itemize}

\section*{Appendix A: IER Algorithm}
In this section, we present the pseudocode of the Instant Episode Repetition (IER) algorithm, as shown in Algorithm~\ref{alg:IER}.

\begin{algorithm}[!htbp]
\small
\caption{Policy Training with Instant Episode Repetition (IER)}
\label{alg:IER}
\begin{algorithmic}[1]
\STATE \textbf{Initialize:} Policy $\pi_\theta$, environment $\mathcal{E}$, buffer $\mathcal{M}$, exploration steps $T_{\text{explore}}$, training steps $T_{\text{train}}$, repetition length $R_N$
\STATE $R_{\text{max}} \gets 0$, $r \gets 0$, $\mathbf{a}^* \gets []$, $s_0 \gets \mathcal{E}.\text{reset}()$, $R_e \gets 0$, $\text{step} \gets 0$
\FOR{$t = 1$ to $T_{\text{train}}$}
    \IF{$t \leq T_{\text{explore}}$}
        \STATE $a_t \sim \mathcal{U}(\mathcal{A})$
    \ELSIF{$r > 0$}
        \STATE $a_t \gets \mathbf{a}^*[\text{step}]$
    \ELSE
        \STATE $a_t \gets \pi_\theta(s_t) + \epsilon_t$
    \ENDIF
    \STATE Execute $a_t$, observe $s_{t+1}, r_t, \text{done}$
    \STATE Store $(s_t, a_t, r_t, s_{t+1}, \text{done})$ in $\mathcal{M}$
    \STATE $R_e \gets R_e + r_t$, $\text{step} \gets \text{step} + 1$
    \STATE Update policy every $K$ steps
    \IF{\text{done or truncated}}
        \IF{$R_e > R_{\text{max}}$}
            \STATE $R_{\text{max}} \gets R_e$, $\mathbf{a}^* \gets$ actions from episode
            \STATE $r \gets R_N$
        \ELSIF{$r > 0$}
            \STATE $r \gets r - 1$
        \ENDIF
        \STATE Reset: $s_0 \gets \mathcal{E}.\text{reset}()$, $R_e \gets 0$, $\text{step} \gets 0$
    \ENDIF
\ENDFOR
\end{algorithmic}
\end{algorithm}

The algorithm begins by initializing the policy, environment, replay buffer, and relevant parameters, such as the number of exploration steps $T_{\text{explore}}$, total training steps $T_{\text{train}}$, and repetition length $R_N$. During the exploration phase, actions are sampled uniformly from the action space to promote state diversity. After the exploration phase, the policy's actions are selected with added exploration noise.

When a new episode achieves a total reward $R_e$ greater than the maximum episode reward $R_{\text{max}}$, the sequence of actions from that episode is stored as $\mathbf{a}^*$. This sequence is then repeated for the next $R_N$ episodes. Repetition ensures that the policy can consistently revisit and learn from the most successful trajectories discovered so far. The variable $r$ tracks the number of remaining repeated episodes.

At each step, the agent interacts with the environment, stores transitions in the replay buffer, and updates the policy at a fixed interval. When an episode terminates, the algorithm checks whether the episode reward exceeds $R_{\text{max}}$ and either updates the best sequence or decrements the repetition counter. This process continues until $T_{\text{train}}$ steps are completed.


\begin{figure}[!htpb]
    \centering
    \begin{subfigure}[b]{0.45\textwidth}
        \includegraphics[width=\textwidth]{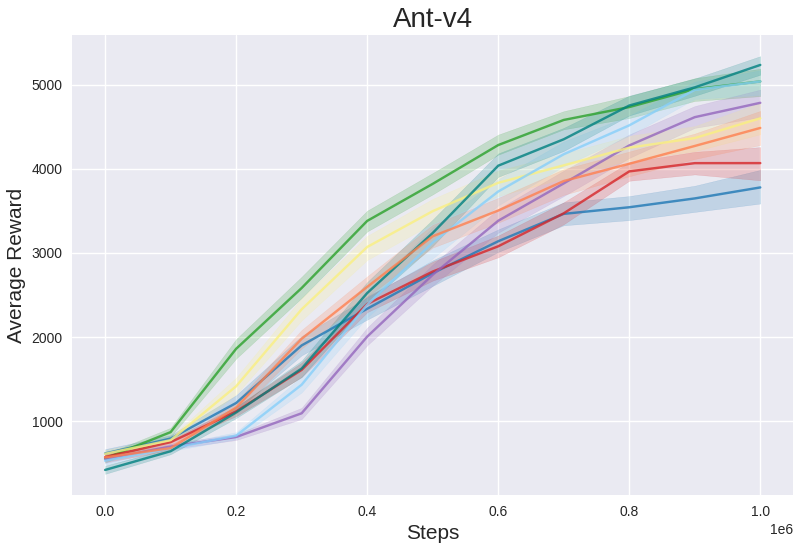}
    \end{subfigure}
    \hfill
    \begin{subfigure}[b]{0.45\textwidth}
        \includegraphics[width=\textwidth]{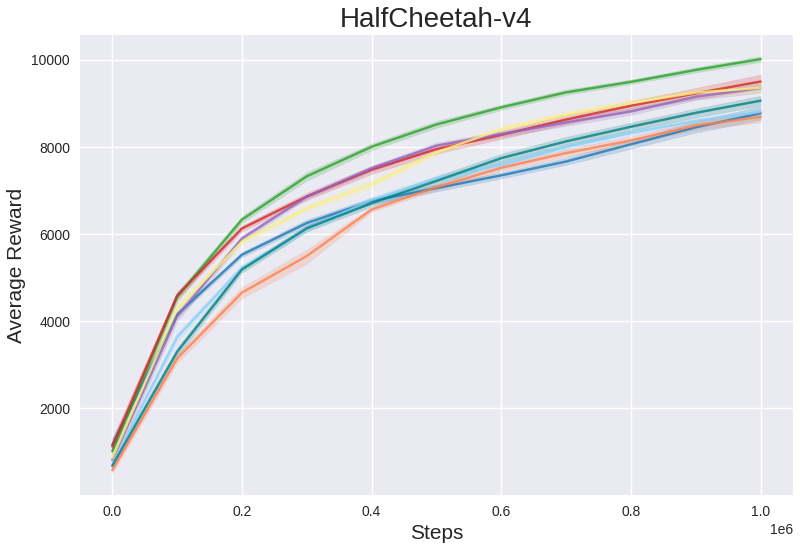}
    \end{subfigure}
    
    \vspace{10pt}
    
    \begin{subfigure}[b]{0.45\textwidth}
        \includegraphics[width=\textwidth]{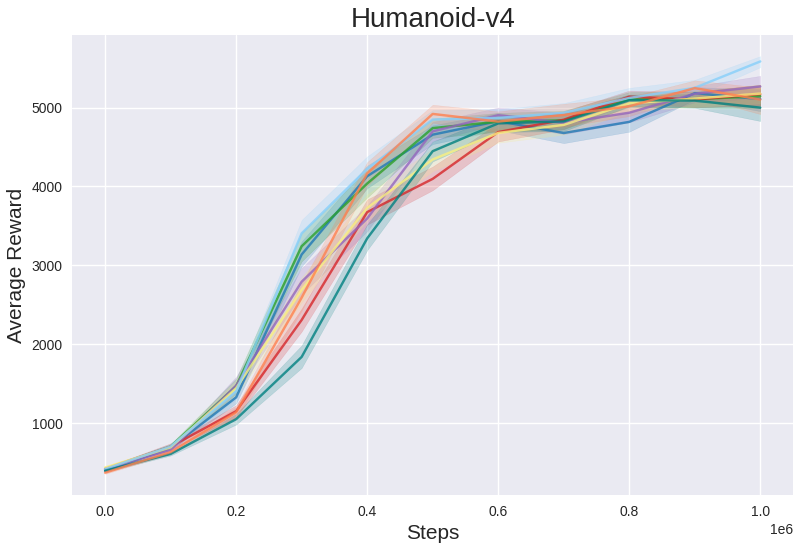}
    \end{subfigure}
    \hfill
    \begin{subfigure}[b]{0.45\textwidth}
        \includegraphics[width=\textwidth]{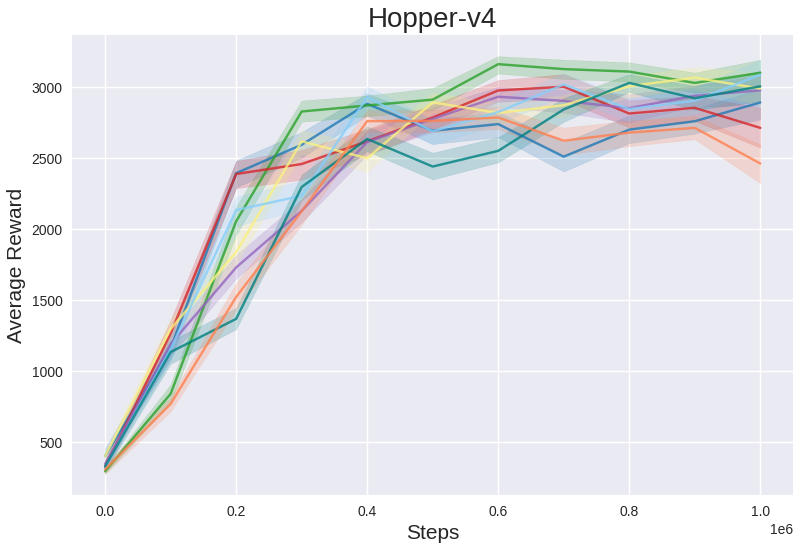}
    \end{subfigure}
    
    \vspace{10pt}
    
    \begin{subfigure}[b]{0.45\textwidth}
        \includegraphics[width=\textwidth]{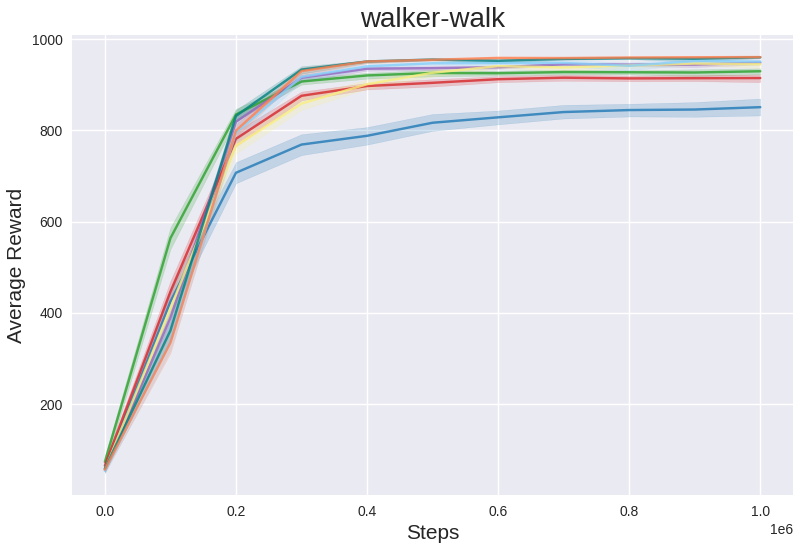}
    \end{subfigure}
    \hfill
    \begin{subfigure}[b]{0.45\textwidth}
        \includegraphics[width=\textwidth]{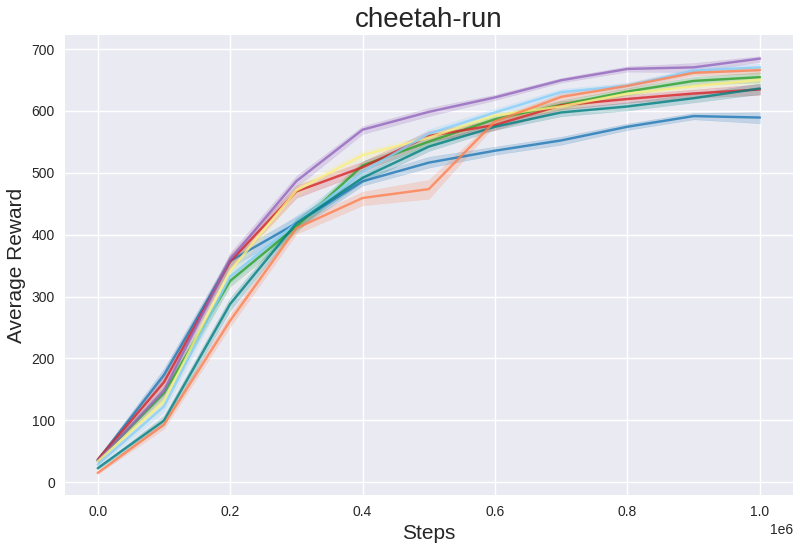}
    \end{subfigure}

    \vspace{10pt}
    
    \begin{subfigure}[b]{0.45\textwidth}
        \includegraphics[width=\textwidth]{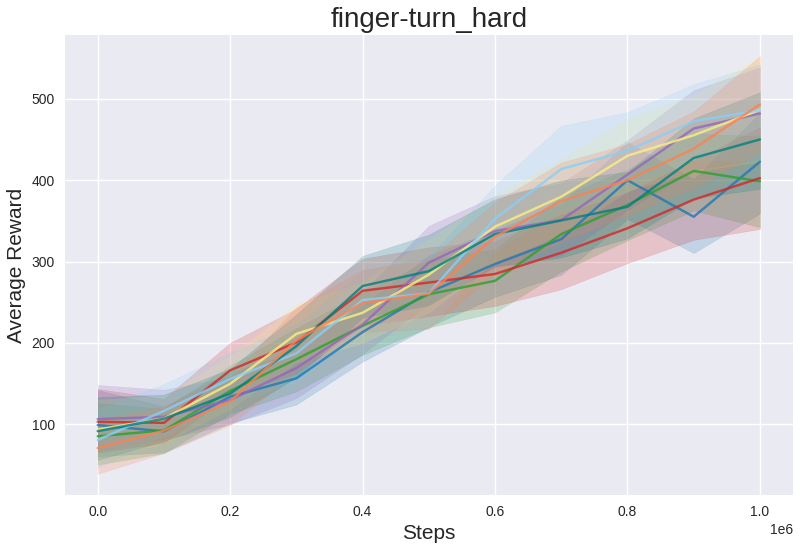}
    \end{subfigure}
    \hfill
    \begin{subfigure}[b]{0.45\textwidth}
        \includegraphics[width=\textwidth]{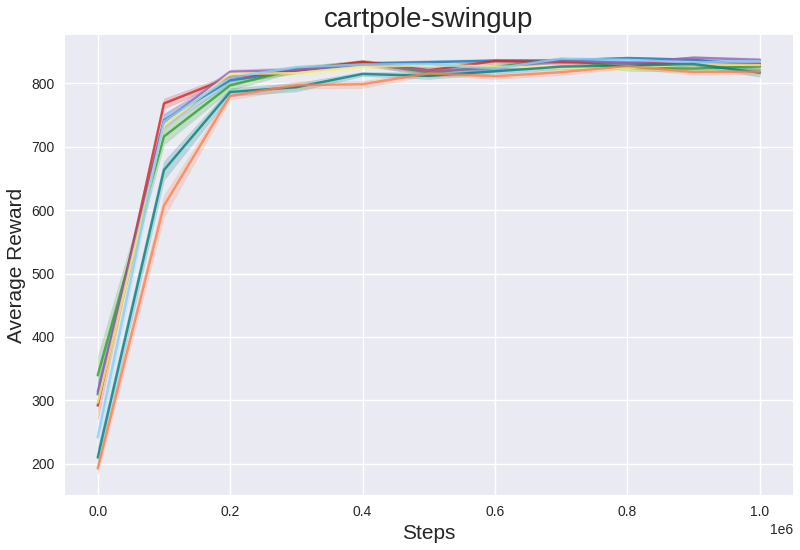}
    \end{subfigure}

    \vspace{5pt}

    \begin{minipage}{0.95\textwidth}
        \centering
        \textcolor{color1}{\rule{0.03\linewidth}{0.2cm}} \scriptsize RN0  \hspace{8pt}
        \textcolor{color2}{\rule{0.03\linewidth}{0.2cm}} \scriptsize RN1 \hspace{8pt}
        \textcolor{color3}{\rule{0.03\linewidth}{0.2cm}} \scriptsize RN2 \hspace{8pt}
        \textcolor{color4}{\rule{0.03\linewidth}{0.2cm}} \scriptsize RN3 \hspace{8pt}
        \textcolor{color5}{\rule{0.03\linewidth}{0.2cm}} \scriptsize RN4 \hspace{8pt}
        \textcolor{color6}{\rule{0.03\linewidth}{0.2cm}} \scriptsize RN5 \hspace{8pt}
        \textcolor{color7}{\rule{0.03\linewidth}{0.2cm}} \scriptsize RN6 \hspace{8pt}
        \textcolor{color8}{\rule{0.03\linewidth}{0.2cm}} \scriptsize RN7
    \end{minipage}
    
    \caption{
       Impact of repetition number RN on \textbf{IER-SAC} performance across all tasks, showing how repeated execution of the highest-reward episode affects learning. Curves show the mean episodic return over ten independent seeds and are smoothed using a sliding window of size 5 for visual clarity.
    }
    \label{fig:IER-SAC-RN}
\end{figure}


\begin{figure}[htpb]
    \centering
    \begin{subfigure}[b]{0.45\textwidth}
        \includegraphics[width=\textwidth]{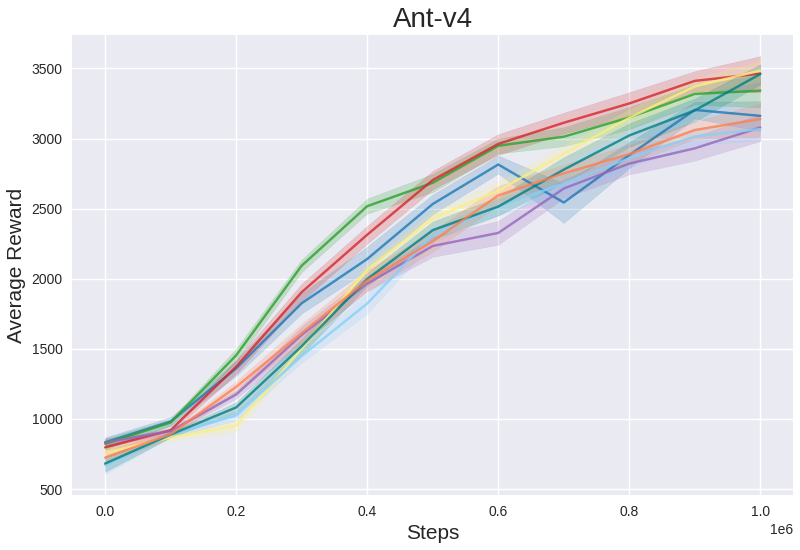}
    \end{subfigure}
    \hfill
    \begin{subfigure}[b]{0.45\textwidth}
        \includegraphics[width=\textwidth]{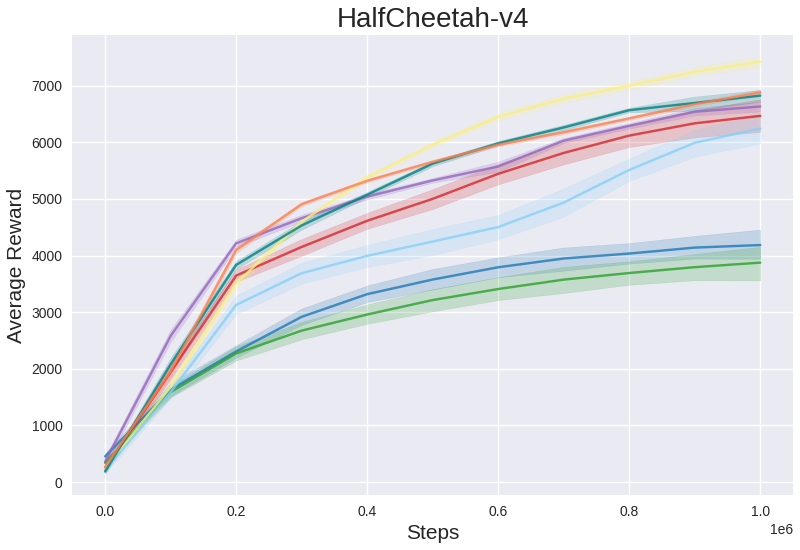}
    \end{subfigure}
    
    \vspace{10pt}
    
    \begin{subfigure}[b]{0.45\textwidth}
        \includegraphics[width=\textwidth]{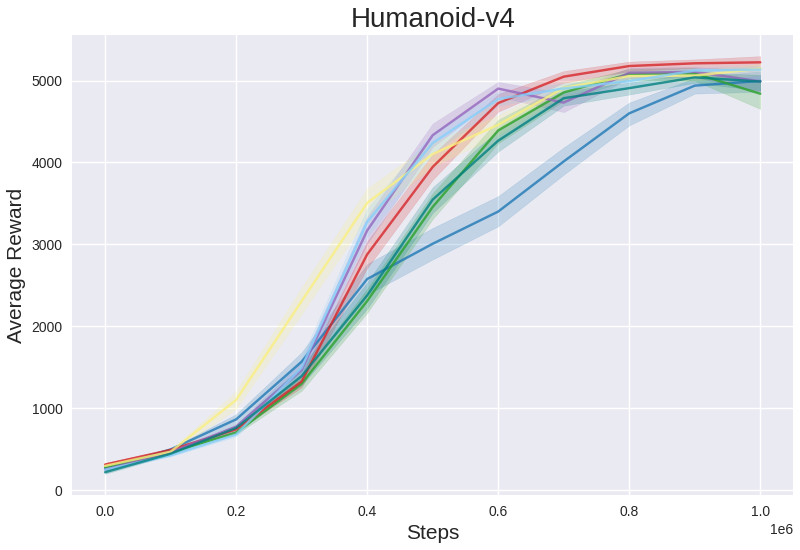}
    \end{subfigure}
    \hfill
       \begin{subfigure}[b]{0.45\textwidth}
        \includegraphics[width=\textwidth]{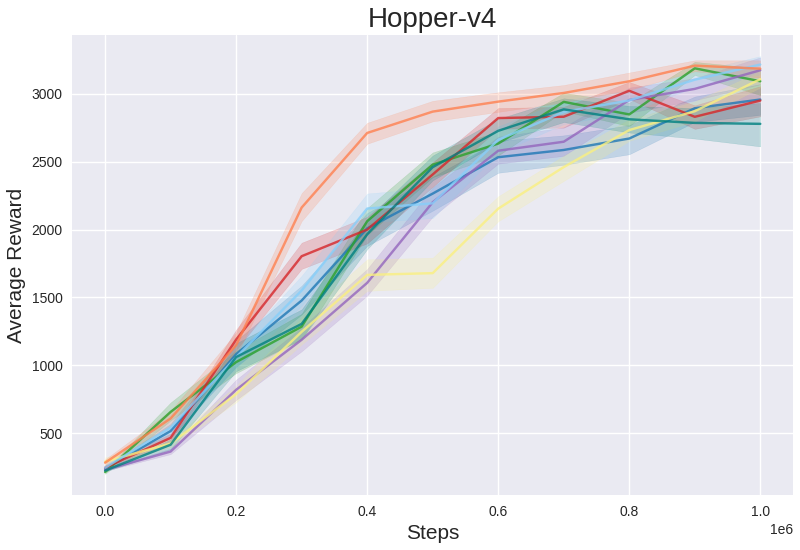}
    \end{subfigure}

    \vspace{10pt}
    
    \begin{subfigure}[b]{0.45\textwidth}
        \includegraphics[width=\textwidth]{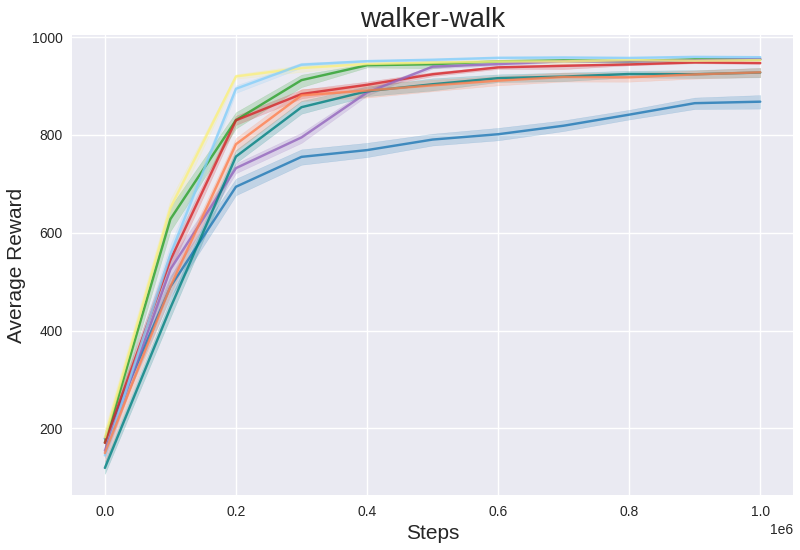}
    \end{subfigure}
     \hfill
    \begin{subfigure}[b]{0.45\textwidth}
        \includegraphics[width=\textwidth]{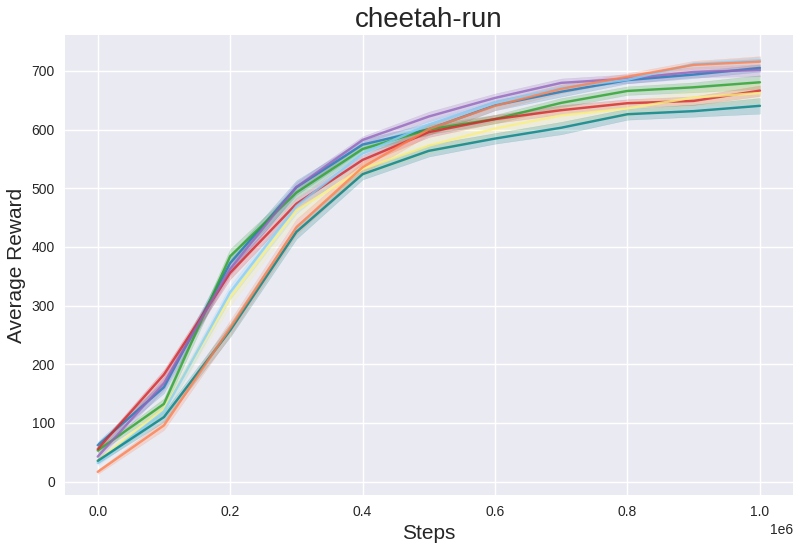}
    \end{subfigure}

    \vspace{10pt}
    
    \begin{subfigure}[b]{0.45\textwidth}
        \includegraphics[width=\textwidth]{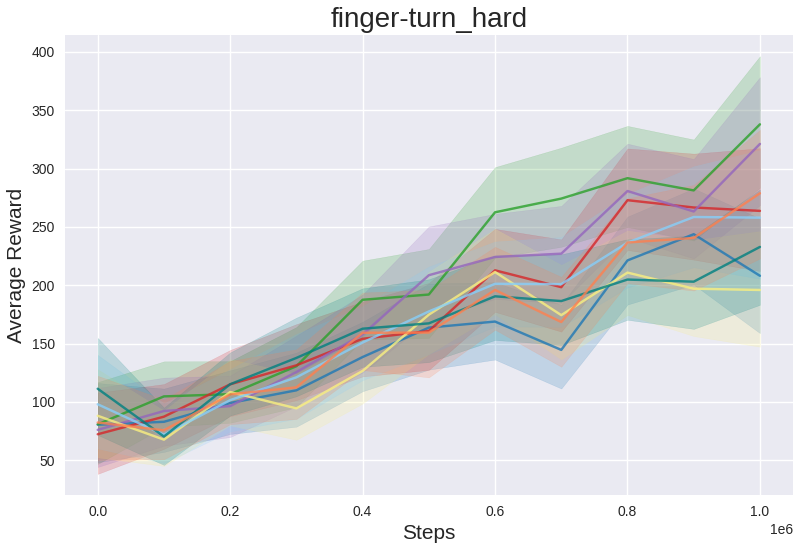}
    \end{subfigure}
    \hfill
    \begin{subfigure}[b]{0.45\textwidth}
        \includegraphics[width=\textwidth]{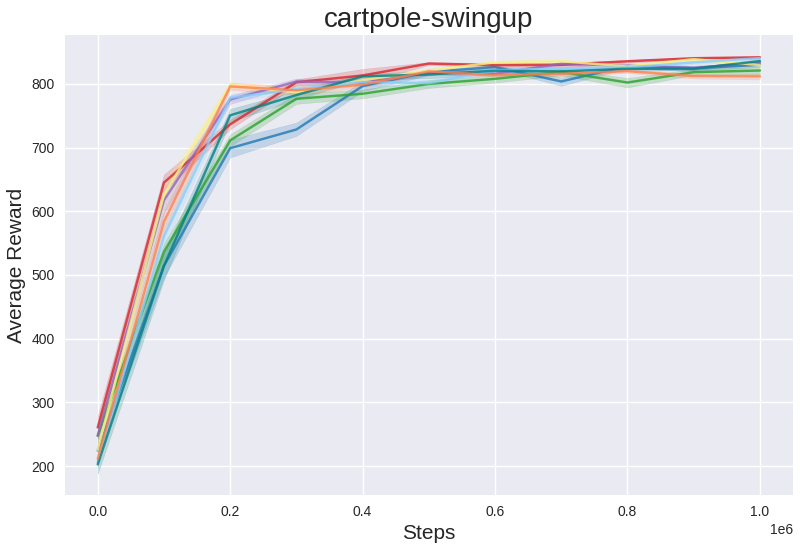}
    \end{subfigure}


    \begin{minipage}{0.95\textwidth}
        \centering
        \textcolor{color1}{\rule{0.03\linewidth}{0.2cm}} \scriptsize RN0  \hspace{8pt}
        \textcolor{color2}{\rule{0.03\linewidth}{0.2cm}} \scriptsize RN1 \hspace{8pt}
        \textcolor{color3}{\rule{0.03\linewidth}{0.2cm}} \scriptsize RN2 \hspace{8pt}
        \textcolor{color4}{\rule{0.03\linewidth}{0.2cm}} \scriptsize RN3 \hspace{8pt}
        \textcolor{color5}{\rule{0.03\linewidth}{0.2cm}} \scriptsize RN4 \hspace{8pt}
        \textcolor{color6}{\rule{0.03\linewidth}{0.2cm}} \scriptsize RN5 \hspace{8pt}
        \textcolor{color7}{\rule{0.03\linewidth}{0.2cm}} \scriptsize RN6 \hspace{8pt}
        \textcolor{color8}{\rule{0.03\linewidth}{0.2cm}} \scriptsize RN7
    \end{minipage}
    
    \caption{
        Impact of repetition number RN on \textbf{IER-TD3} performance across all tasks, showing how repeated execution of the highest-reward episode affects learning. Curves show the mean episodic return over ten independent seeds and are smoothed using a sliding window of size 5 for visual clarity.
    }
    \label{fig:IER-TD3-RN}
\end{figure}

\begin{figure}[!htbp]
    \centering
    \begin{subfigure}[t]{0.24\textwidth}%
        \centering
        \includegraphics[width=\linewidth]{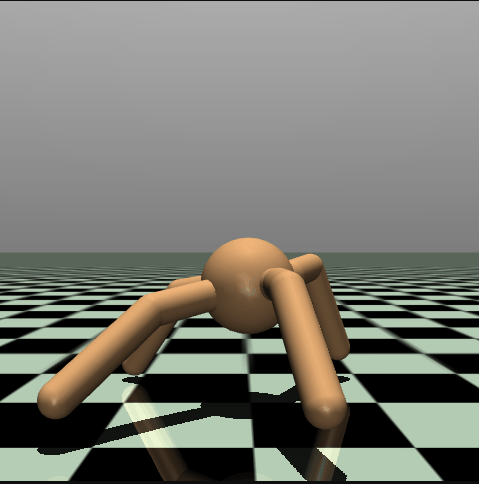}%
        \caption{Ant-v4}%
        \label{fig:ant}%
    \end{subfigure}%
    \begin{subfigure}[t]{0.24\textwidth}%
        \centering
        \includegraphics[width=\linewidth]{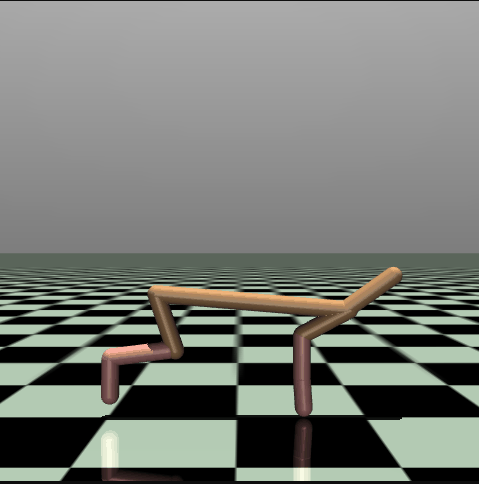}%
        \caption{HalfCheetah-v4}%
        \label{fig:halfcheetah}%
    \end{subfigure}%
    \begin{subfigure}[t]{0.24\textwidth}%
        \centering
        \includegraphics[width=\linewidth]{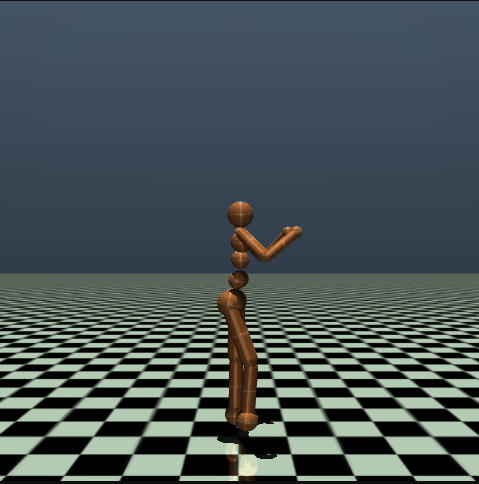}%
        \caption{Humanoid-v4}%
        \label{fig:humanoid}%
    \end{subfigure}%
    \begin{subfigure}[t]{0.243\textwidth}%
        \centering
        \includegraphics[width=\linewidth]{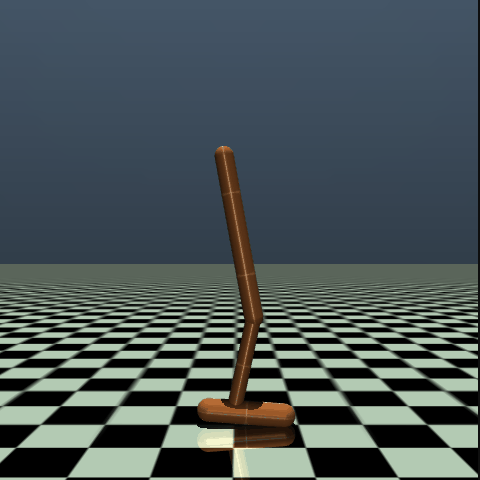}%
        \caption{Hopper-v4}%
        \label{fig:hopper}%
    \end{subfigure}%

    \vspace{10pt}
    
    \begin{subfigure}[t]{0.24\textwidth}%
        \centering
        \includegraphics[width=\linewidth]{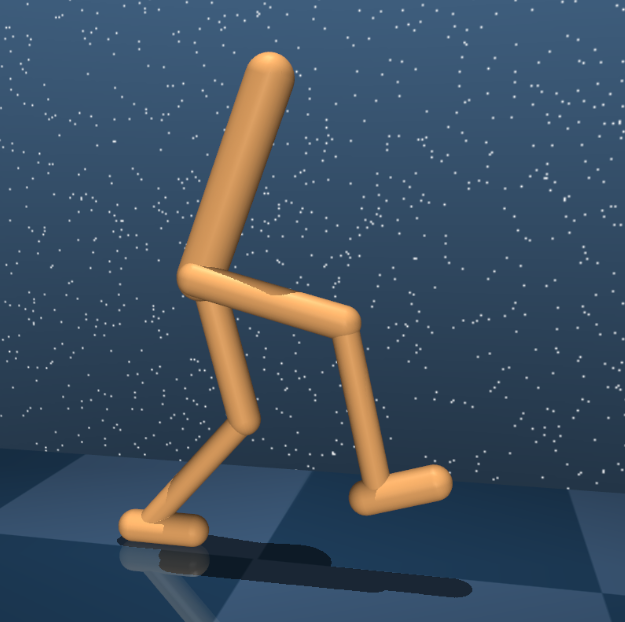}%
        \caption{Walker-Walk}%
        \label{fig:walker}%
    \end{subfigure}%
    \begin{subfigure}[t]{0.24\textwidth}%
        \centering
        \includegraphics[width=\linewidth]{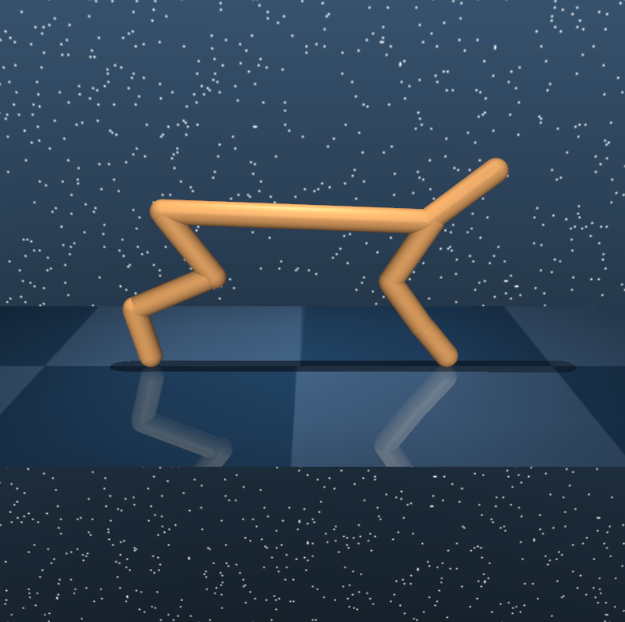}%
        \caption{Cheetah-Run}%
        \label{fig:cheetah}%
    \end{subfigure}%
    \begin{subfigure}[t]{0.24\textwidth}%
        \centering
        \includegraphics[width=\linewidth]{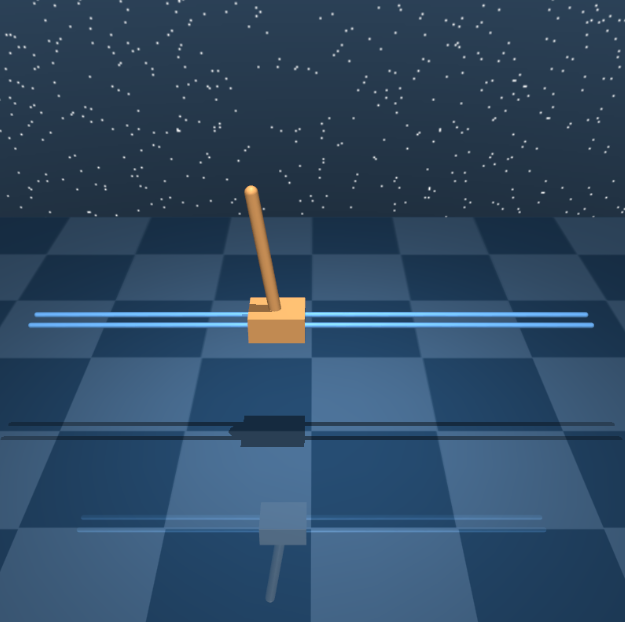}%
        \caption{Cartpole-Swingup}%
        \label{fig:cartpole}%
    \end{subfigure}%
    \begin{subfigure}[t]{0.244\textwidth}%
        \centering
        \includegraphics[width=\linewidth]{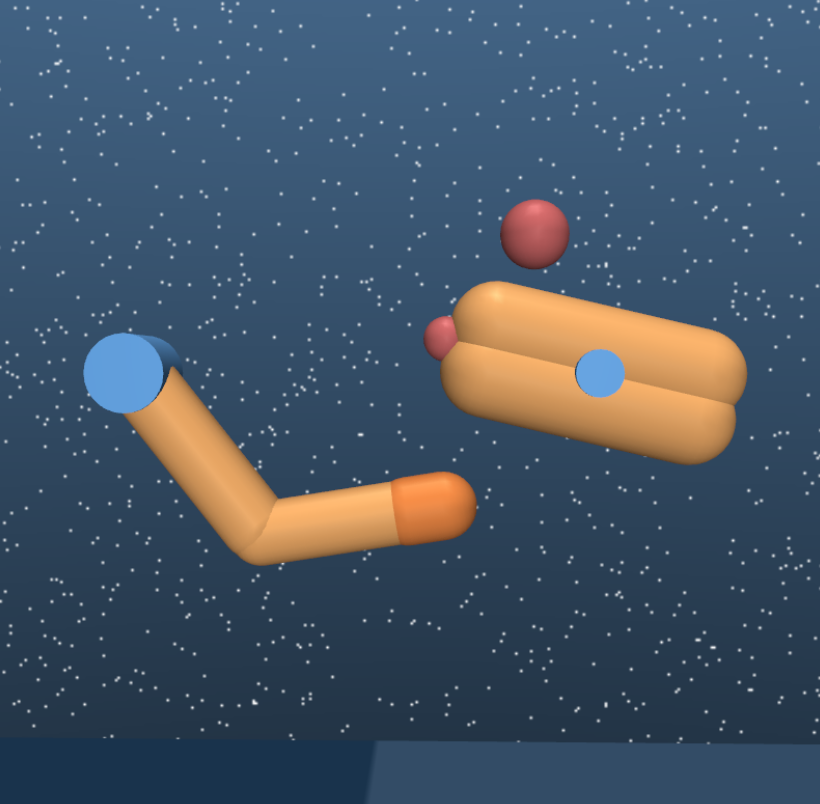}%
        \caption{Finger-Turn-Hard}%
        \label{fig:finger}%
    \end{subfigure}%

    \caption{tasks used in our experiments.}
    \label{fig:mujoco_envs}
\end{figure}

\section*{Appendix B: Extended Analysis of the Repetition Number (RN)}

This appendix provides additional experimental results analyzing the
 impact of the RN used in the IER algorithm on agent performance. 

Figures~\ref{fig:IER-SAC-RN} and~\ref{fig:IER-TD3-RN} present complete performance curves across all RN values and tasks, allowing for the comparison of performance trends under varying repetition frequencies.

Across nearly all tasks, RN=0 consistently results in one of the weakest performances, confirming that learning without repetition underutilizes successful trajectories. Even a small amount of repetition (e.g., RN=1) consistently improves performance, indicating that revisiting high-reward episodes enhances learning efficiency. Higher RNs, such as RN=6, continue to provide strong performance in many tasks; however, performance occasionally declines at RN=7, likely due to overexploitation or reduced diversity in the sampled experiences.

\paragraph{Domain-Level Analysis.}

\begin{figure}[!htbp]
    \centering
    
    \begin{subfigure}[t]{0.49\textwidth}
        \centering
        \includegraphics[width=\textwidth]
        {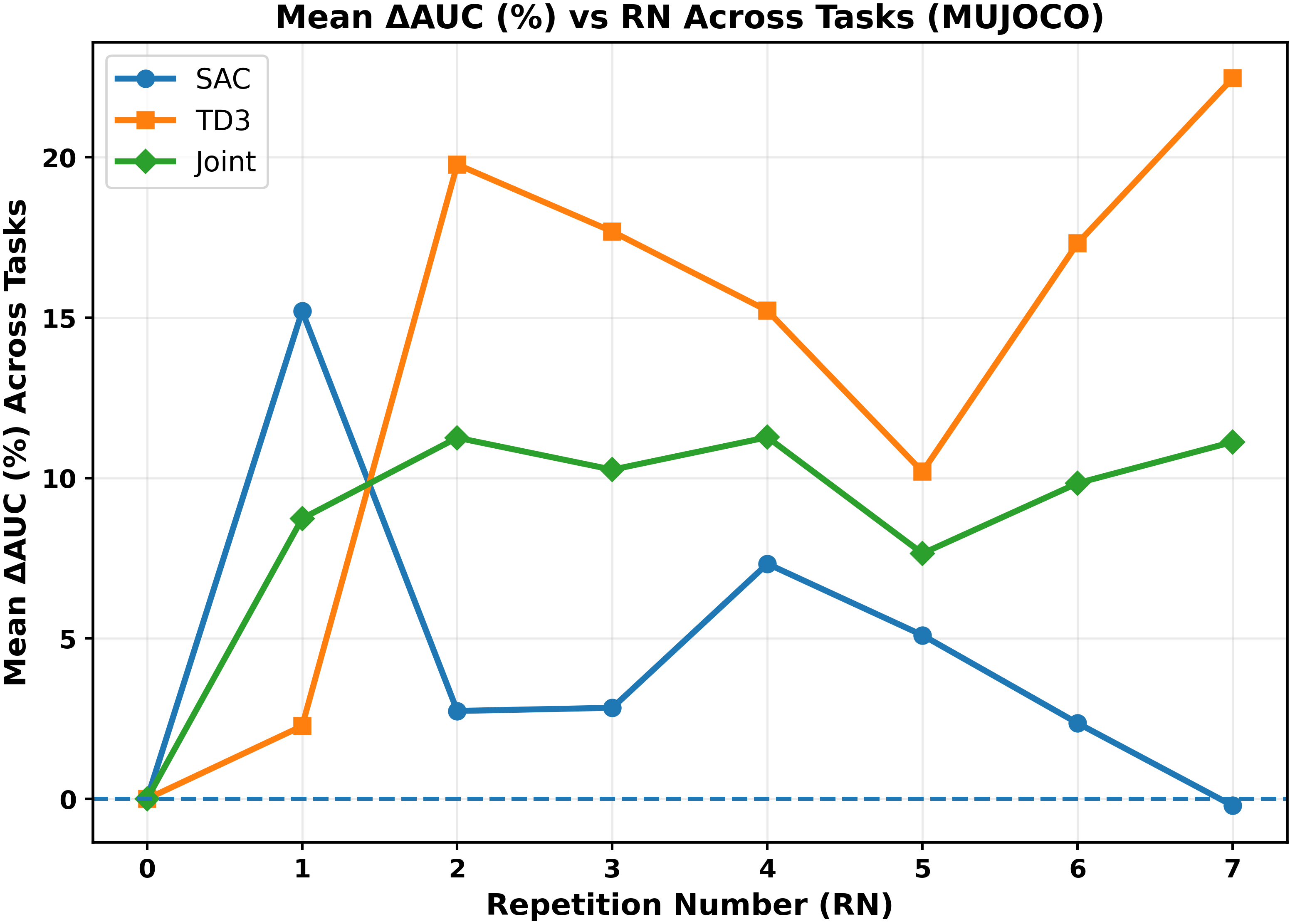}
        \caption{Mean $\Delta$AUC\% across MuJoCo tasks. Moderate repetition (RN3--RN4) yields the strongest gains, particularly under TD3.}
        \label{fig:mean_delta_curve_mujoco}
    \end{subfigure}
    \hfill
    \begin{subfigure}[t]{0.49\textwidth}
        \centering
        \includegraphics[width=\textwidth]
        {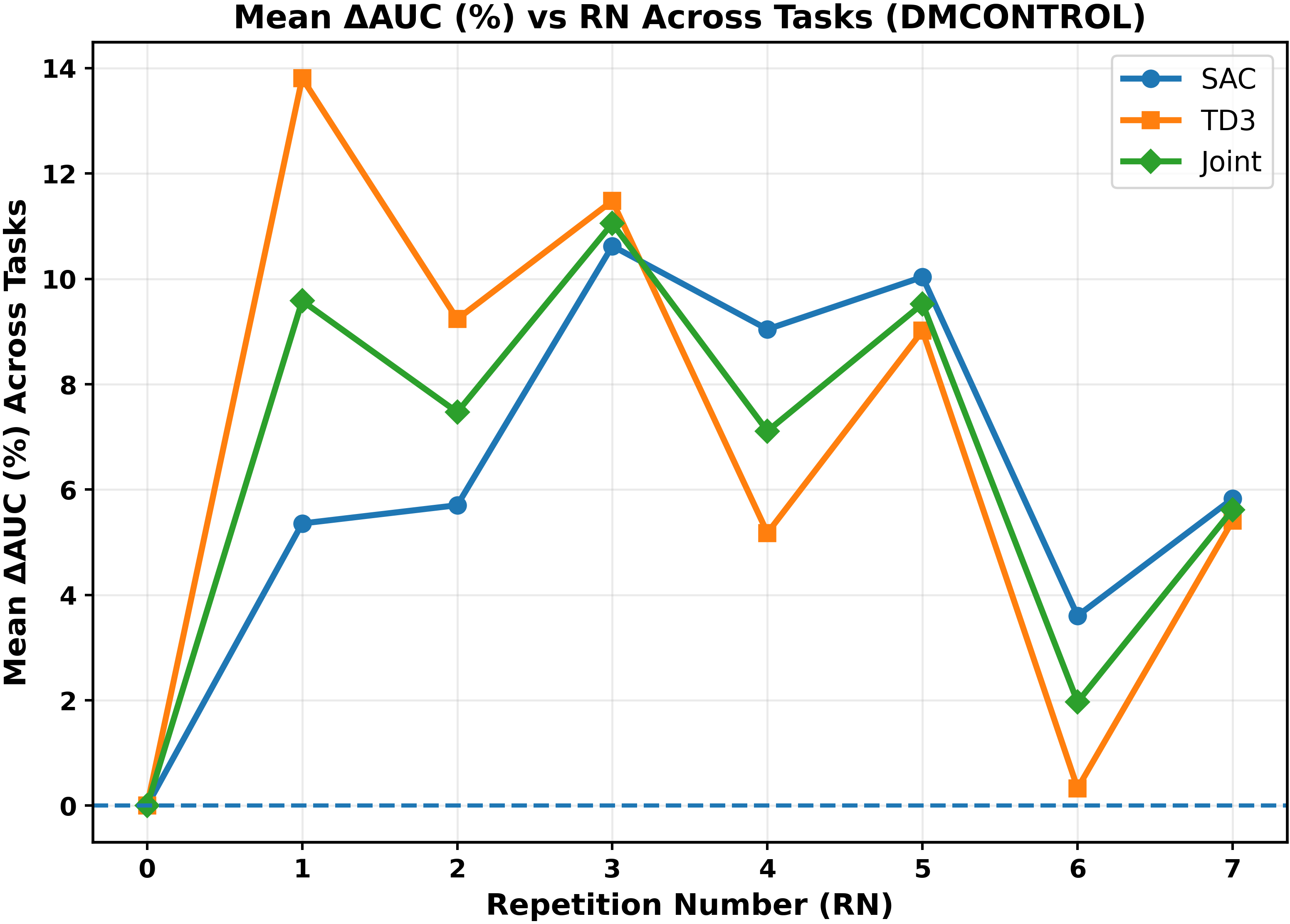}
        \caption{Mean $\Delta$AUC\% across DeepMind Control tasks. Improvements are more uniform and less sensitive to RN magnitude.}
        \label{fig:mean_delta_curve_dmcontrol}
    \end{subfigure}

    \caption{Mean $\Delta$AUC\% across environment groups. Left: MuJoCo tasks. Right: DeepMind Control tasks.}
    \label{fig:mean_delta_curve_combined}
\end{figure}

Figures~\ref{fig:mean_delta_curve_mujoco} and~\ref{fig:mean_delta_curve_dmcontrol} separate MuJoCo and DeepMind Control tasks.

In MuJoCo locomotion environments, improvements are greatest at moderate repetition levels, particularly under TD3. In DeepMind Control tasks, repetition effects are smoother and less sensitive to RN magnitude. This suggests that repetition interacts with environment dynamics, with higher-dimensional locomotion tasks benefiting more strongly from moderate replay.

\paragraph{Agreement Between Baseline Algorithms.}

To evaluate robustness beyond single-task optima, we first identify, for each task and each baseline (IER-SAC and IER-TD3), the set of RN values whose normalized AUC is within 95\% of that task’s best AUC. We refer to these as the near-optimal RN sets.

An agreement is defined as selecting an RN who performs well for both baselines on the same task. When the near-optimal RN sets of SAC and TD3 overlap, we choose the RN from their intersection, i.e., an RN that is simultaneously near-optimal for both algorithms.

If no intersection exists, we apply a rank-based criterion. Specifically, RN values are ranked according to their normalized AUC under each baseline, and we select the RN with the best combined rank across SAC and TD3, ensuring a balanced compromise rather than favoring a single algorithm (Table~\ref{tab:agreement_rn}).

\begin{figure}[!htbp]
    \centering
    \includegraphics[width=0.5\textwidth]{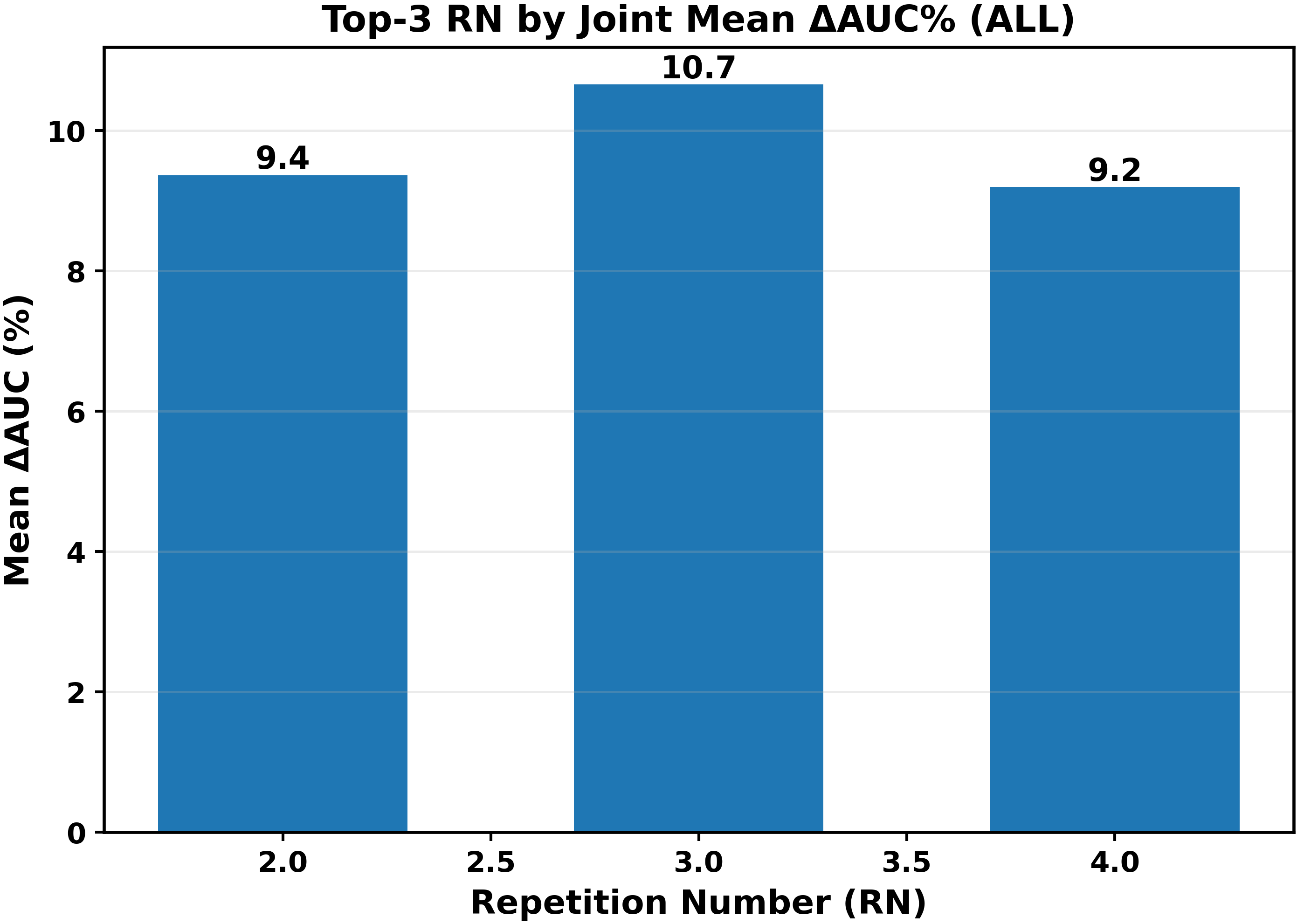}
    \caption{
    Frequency of RN values selected as robust joint candidates across tasks. RN2, RN3, and RN5 appear most frequently.
    }
    \label{fig:rn_frequency}
\end{figure}

Several tasks exhibit direct intersection-based agreement, while others require rank-based compromise selection. Figure~\ref{fig:rn_frequency} summarizes the frequency of selected agreement RN values across tasks. RN2, RN3, and RN5 appear most frequently, indicating that moderate repetition levels tend to generalize across algorithms.

The results suggest that moderate repetition provides the most reliable performance across tasks and baseline algorithms. Very small RN values underutilize successful trajectories, whereas excessively large RN values yield diminishing returns. Across MuJoCo and DeepMind Control environments, RN2--RN4 offers a stable and effective operating range for IER.

\begin{table}[!htbp]
\centering
\caption{Per-task RN agreement based on 95\% near-best criterion.}
\label{tab:agreement_rn}
\begin{tabular}{lcc}
\hline
\textbf{Task} & \textbf{Agreement RN} & \textbf{Selection Type} \\
\hline
Ant-v4 & RN1 & Intersection \\
Cheetah-Run & RN3 & Intersection \\
Walker-Walk & RN5 & Intersection \\
Cartpole-Swingup & RN2 & Intersection \\
HalfCheetah-v4 & RN4 & Rank-based \\
Hopper-v4 & RN2 & Rank-based \\
Humanoid-v4 & RN3 & Rank-based \\
Finger-Turn-Hard & RN5 & Rank-based \\

\hline
\end{tabular}
\end{table}

\section*{Appendix C: Environment Descriptions}

\begin{table}[!htbp]
    \centering
    \caption{Summary of environments used in evaluation. }
    \label{tab:env-summary}
    \begin{tabular}{p{3.5cm} p{2.2cm} p{2.5cm} p{4.2cm} p{2.2cm}}
        \toprule
        \textbf{Environment} & \textbf{Suite} & \textbf{Task Type} & \textbf{Description} & \textbf{Approx. Difficulty} \\
        \midrule
        Ant-v4 & MuJoCo & Locomotion & Quadruped; coordination-heavy with complex contact dynamics. & High \\
        HalfCheetah-v4 & MuJoCo & Locomotion & Planar biped; dense unbounded rewards support long-term learning. & Medium \\
        Humanoid-v4 & MuJoCo & Locomotion & Biped with high DOF; unstable dynamics and sensitive to control noise. & High \\
        Hopper-v4 & MuJoCo & Locomotion & Single-legged hopping; requires balance and precise control, with early termination on failure. & Medium \\
        Walker-Walk & DMC & Locomotion & Bipedal walking; fewer DOF than Humanoid, but still unstable. & Medium \\
        Cheetah-Run & DMC & Locomotion & Fast planar cheetah; velocity-based control under bounded rewards. & Low \\
        Cartpole-Swingup & DMC & Classic Control & Swing-up and balance pole on cart; sparse rewards, underactuated. & Medium \\
        Finger-TurnHard & DMC & Manipulation & Rotate object with a single finger; fine-grained actuation required. & High \\

        \bottomrule
    \end{tabular}
\end{table}
\begin{table}[!htbp]
    \centering
    \caption{Observation and action space dimensions and episode horizon for each environment. $\mathbb{R}^d$ denotes a $d$-dimensional real vector; $[-1, 1]^d$ denotes a bounded action space.}
    \label{tab:env-dimensions}
    \begin{tabular}{lccc}
        \toprule
        \textbf{Environment} & \textbf{Observation Space} & \textbf{Action Space} & \textbf{Horizon} \\
        \midrule
        \multicolumn{4}{c}{\textit{MuJoCo (vector-based)}} \\
        \midrule
        Ant-v4         & $\mathbb{R}^{111}$ & $[-1, 1]^8$    & 1000 (may terminate early due to instability) \\
        HalfCheetah-v4 & $\mathbb{R}^{17}$  & $[-1, 1]^6$    & 1000 \\
        Humanoid-v4    & $\mathbb{R}^{376}$ & $[-1, 1]^{17}$ & 1000 (may terminate early due to instability) \\
        Hopper-v4      & $\mathbb{R}^{11}$  & $[-1, 1]^3$    & 1000 \\

        \midrule
        \multicolumn{4}{c}{\textit{DeepMind Control Suite (vector-based)}} \\
        \midrule
        
        Walker-Walk        & $\mathbb{R}^{24}$ & $[-1, 1]^6$ & 1000 \\
        Cheetah-Run        & $\mathbb{R}^{17}$ & $[-1, 1]^6$ & 1000 \\
        Cartpole-Swingup   & $\mathbb{R}^{5}$  & $[-1, 1]^1$ & 1000 \\
        Finger-TurnHard    & $\mathbb{R}^{12}$ & $[-1, 1]^2$ & 1000 \\
        \bottomrule
    \end{tabular}
\end{table}

We evaluate our approach across a diverse set of continuous control environments from MuJoCo~\cite{todorov2012mujoco} and the DeepMind Control Suite (DMC)~\cite{tassa2018deepmind}, two widely adopted platforms for studying motor control in physically realistic simulations. All tasks are configured with low-dimensional, vector-based observations consisting of joint positions, velocities, orientations, and other proprioceptive signals—excluding high-dimensional visual input.
\paragraph{MuJoCo Environments.}
MuJoCo tasks, accessed via the Gym interface, focus on high-speed and dynamic locomotion with varying complexity:

\begin{itemize}

    \item \textbf{Ant-v4:} A quadrupedal agent learns to walk using a high-dimensional action space. Coordination and contact dynamics make this task nontrivial.
    
    \item \textbf{HalfCheetah-v4:} A planar biped with a flexible spine learns to run. It offers dense, unbounded rewards, with cumulative rewards often exceeding 12,000, making it suitable for evaluating long-horizon learning and experience prioritization.
    
    \item \textbf{Humanoid-v4:} A high-DOF biped must learn to walk upright. Its instability and dimensionality pose significant challenges for balance, control, and exploration.
    
    \item \textbf{Hopper-v4:} A single-legged agent must learn to hop forward while maintaining balance. The task requires precise control and stability, as failure to maintain upright posture results in early episode termination. Its sensitivity to instability makes it useful for assessing learning robustness.
\end{itemize}

\begin{table}[!htbp]
    \centering
    \caption{Real-World Robotic Gripper Task Characteristics}
    \label{tab:real-world-task}
    \begin{tabular}{p{4.5cm} p{8.5cm}}
        \toprule
        \textbf{Feature} & \textbf{Description} \\
        \midrule
        \multicolumn{2}{l}{\textbf{Task Overview}} \\
        Task Type & Dynamic cube translation with robotic gripper \\
        Object & Cube tracked via ArUco markers \\
        Goal & Translate cube as far as possible within camera view \\
        Reward Components & Force-closure grasp + lateral displacement \\
        \midrule
        \multicolumn{2}{l}{\textbf{Hardware Setup}} \\
        Gripper Dexterity & Stable force closure; no extrinsic dexterity or caging \\
        Workspace & Camera-bounded area (blue rectangle, Fig.~\ref{fig:gripper}) \\
        Materials & Consumer-grade aluminium and 3D-printed parts \\
        Sensors & Logitech webcam with ArUco tracking \\
        Actuation & Dynamixel servos \\
        \midrule
        \multicolumn{2}{l}{\textbf{Control and Learning}} \\
        State Space & Servo angles; cube current/previous positions (x, y mm) \\
        Action Space & Target servo joint angles \\
        Episode Length & 20 real-world steps \\
        Training Steps & 20,000 \\
        Reset Mechanism & Rack-and-pinion elevator for automatic repositioning \\
        \midrule
        \multicolumn{2}{l}{\textbf{System Specifications}} \\
        PC Specs & Intel i9-10900, 128GB RAM, Nvidia RTX 3080 \\
        \bottomrule
    \end{tabular}
\end{table}

\paragraph{DeepMind Control Suite Environments.}
DMC tasks are designed for general motor control using structured state vectors. Unlike MuJoCo, rewards are typically bounded in $[0, 1000]$, focusing more on precision than accumulation:

\begin{itemize}

    \item \textbf{Walker-Walk:} A biped learns forward locomotion while maintaining balance. It involves moderately complex dynamics and fewer degrees of freedom than Humanoid.
    
    \item \textbf{Cheetah-Run:} A planar cheetah learns forward locomotion under aggressive dynamics and bounded rewards. Although structurally similar to HalfCheetah-v4, it exhibits greater control sensitivity and performance saturation. We deliberately include this task to evaluate robustness and generalization in environments with similar morphology but different reward scales and control challenges.
    
    \item \textbf{Cartpole-Swingup:} An underactuated classic control problem requiring swing-up and stabilization of a pole mounted on a cart. Sparse rewards and nonlinear dynamics make learning nontrivial.
    
    \item \textbf{Finger-TurnHard:} A single robotic finger must rotate an object to a target orientation. Unlike the easy variant, it requires more precise torque control and sustained actuation under stricter success conditions.

\end{itemize}

\begin{figure}[!htbp]
    \centering
    \includegraphics[width=0.45\linewidth]{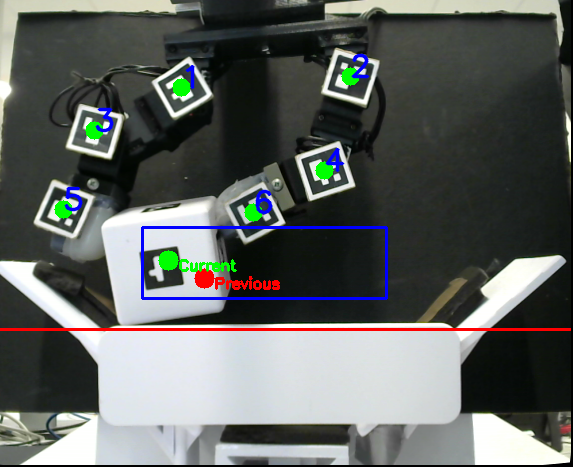}
    \caption[Workspace and cube tracking]{Workspace boundaries and cube tracking for reward evaluation. Green and red markers indicate current and previous cube positions used to compute displacement-based rewards.}
    \label{fig:gripper}
   \end{figure}

\renewcommand{\arraystretch}{0.9}

\begin{table}[!t]
\centering
\caption{Hyperparameters used in training. Shared parameters apply to both SAC and TD3; SAC, TD3, and SIL sections list algorithm-specific settings.}
\label{tab:hyperparameters}
\begin{tabular}{lp{0.78\textwidth}}
\hline
\textbf{Parameter} & \textbf{Value (Description)} \\
\hline
\multicolumn{2}{c}{\textbf{Shared Parameters}} \\
batch\_size & 256 (Mini-batch size) \\
buffer\_size & $10^6$ (Replay buffer size) \\
max\_steps\_exploration & $10^3$ (Exploration steps before training) \\
max\_steps\_training & $10^6$ (Total training steps) \\
number\_steps\_per\_evaluation & $10^4$ (Steps between evaluations) \\
number\_eval\_episodes & 10 (Episodes per evaluation) \\
number\_steps\_per\_train\_policy & 1 (Steps per policy update) \\
min\_noise & 0.0 (Min exploration noise) \\
noise\_scale & 0.1 (Initial Gaussian noise scale) \\
noise\_decay & 1.0 (Exploration noise decay) \\
actor\_lr & 3e-4 (Actor learning rate) \\
critic\_lr & 3e-4 (Critic learning rate) \\
gamma & 0.99 (Discount factor) \\
tau & 0.005 (Target smoothing) \\
hidden\_size\_actor & [256, 256] (Actor hidden layers) \\
hidden\_size\_critic & [256, 256] (Critic hidden layers) \\
optimizer & Adam (Training optimizer) \\
nonlinearity & ReLU (Network activation) \\
\hline
\multicolumn{2}{c}{\textbf{TD3-Specific}} \\
policy\_update\_freq & 2 (Policy update frequency) \\
\hline
\multicolumn{2}{c}{\textbf{SAC-Specific}} \\
alpha\_lr & 3e-4 (Entropy temperature LR) \\
reward\_scale & 1.0 (Reward scaling) \\
log\_std\_bounds & [-20, 2] (Log-std bounds) \\
policy\_update\_freq & 1 (Policy update frequency) \\
target\_update\_freq & 1 (Target update frequency) \\
\hline
\multicolumn{2}{c}{\textbf{SIL-Specific}} \\
initial\_per\_exponents & (0.7, 0.4) (Initial PER exponents $(\alpha, \beta)$) \\
replay\_period & 64 (Replay sampling frequency) \\
gradient\_step & 64 (Gradient updates per sampling) \\
max\_nlogp & 5 (Max negative log probability) \\
\hline
\end{tabular}
\end{table}

\subsection*{Real-World Robotic Task}

This section describes the Dynamic Object Translation task involving gripper-based manipulation, including the hardware setup, sensing modalities, and key challenges.

The task, illustrated in Fig.~\ref{fig:gripper}, is a dexterous in-hand manipulation challenge requiring the gripper to securely grasp a cube and manipulate it to a desired position within image space. Unlike tasks leveraging extrinsic dexterity \cite{dafle2014extrinsic} or caging strategies, this setup demands sustained force closure to maintain grasp stability throughout the manipulation.

This task builds upon a prior two-finger push manipulation setup from \cite{valencia2024magebaseddeepreinforcementlearning}, with a 90-degree rotation along the x-axis that orients the gripper downward. This modification eliminates the supporting base beneath the cube and intensifies gravitational effects, forcing the gripper to develop stable grasp and finger manipulation strategies to avoid dropping the object.

Success in this task is defined by continuously manipulating the cube, maximizing its displacement while maintaining a stable grasp. The task pushes beyond simulation by involving intrinsic gripper dexterity in a physical environment, introducing real-world challenges such as limited data collection and hardware constraints.

Fig.~\ref{fig:gripper} shows the workspace from a camera viewpoint. The blue rectangle denotes the gripper's operational area. The cube's current and previous positions are tracked using ArUco markers, visualized as green and red circles, respectively, providing intuitive feedback for monitoring and debugging. The hardware combines aluminum extrusions and 3D-printed joints, with a Logitech webcam capturing the scene. Experiments run on a single PC, ensuring real-time control. The gripper hardware is built with consumer-level components and extensive 3D printing, yielding a cost-effective setup. An automated reset mechanism is integrated to facilitate efficient data collection and consistent training. A rack-and-pinion elevator lifts the cube to a fixed start position between the gripper's fingertips at the end of each episode, enabling rapid and repeatable task restarts.
A detailed summary of the task’s characteristics is provided in Table~\ref{tab:real-world-task}.

\paragraph{Reward Structure:}  
Rewards are computed based on two main criteria: (1) successful force-closure grasping of the object, and (2) the magnitude of lateral displacement of the cube between consecutive time steps. This reward structure encourages the agent to maintain grip stability while steadily moving the object over longer distances.

\paragraph{State and Action Spaces:}  
The \textbf{state space} is a vector consisting of the positions of the servos, Aruco markers in the image (xy), and cube positions (xy). Similarly, the action space is represented as a vector containing the desired positions of each servo joint. Each position value in this vector corresponds to a specific angular position for the Dynamixel servos. The vector's elements directly control the orientation and configuration of the gripper, allowing for precise manipulation of objects.

\paragraph{Training Setup:}  
Agents are trained for 20,000 interaction steps. Each episode comprises 20 real-time steps, reflecting physical environment constraints.

\section*{Appendix D: Configuration and Hyperparameter Settings}

This appendix provides the complete configuration and hyperparameter settings used in all experiments. All agents were implemented in PyTorch, following architectures and optimization settings consistent with prior work~\cite{fujimoto2018addressing, oh2021model}. To ensure fair comparisons, all tasks were trained under the same configuration unless specified otherwise. Experiments in the main study were averaged over five random seeds, while the ablation study on RN used ten seeds to improve statistical reliability.

Table~\ref{tab:hyperparameters} summarizes the hyperparameters used in training. Shared parameters apply to both SAC and TD3, while algorithm-specific settings for SAC, TD3, and the SIL extension are listed separately. These configurations were kept consistent across tasks to isolate the effects of algorithmic modifications, such as varying the RN or incorporating intrinsic motivation mechanisms.

\end{document}